\documentclass[10pt,twocolumn,letterpaper]{article}

\usepackage[table]{xcolor}
\usepackage[pagenumbers]{iccv} % To force page numbers, e.g. for an arXiv version
\usepackage{epsfig}
\usepackage{graphicx}
\usepackage{amsmath}
\usepackage{amssymb}
\usepackage{booktabs}
\usepackage{bm}
\usepackage{amsfonts}
\usepackage{mathrsfs}
\usepackage{pifont}
\usepackage{float}
\usepackage{subcaption}
\usepackage{indentfirst}
\usepackage{algorithm}
\usepackage{bbding}
\usepackage{multicol}
\usepackage{url}
\usepackage{enumitem}
\usepackage{algpseudocode}

\definecolor{iccvblue}{rgb}{0.21,0.49,0.74}
\definecolor{iccvgreen}{rgb}{0.56,0.93,0.56}
\usepackage[pagebackref=true,breaklinks=true,letterpaper=true,colorlinks,bookmarks=false,citecolor=iccvgreen]{hyperref}

\definecolor{SoftRed}{RGB}{255,102,102}

\usepackage[capitalize]{cleveref}
\crefname{section}{Sec.}{Secs.}
\Crefname{section}{Section}{Sections}
\Crefname{table}{Table}{Tables}
\crefname{table}{Tab.}{Tabs.}

\usepackage{xspace}

\def\paperID{****} % *** Enter the Paper ID here
\def\confName{ICCV}
\def\confYear{2025}

\title{{\color{SoftRed}{$\mathbf{R^3}$}}-Avatar: {\color{SoftRed}{$\mathbf{R}$}}ecord and {\color{SoftRed}{$\mathbf{R}$}}etrieve Temporal Codebook for {\color{SoftRed}{$\mathbf{R}$}}econstructing Photorealistic Human Avatars}

\author{
    Yifan Zhan\textsuperscript{1,2}\footnotemark[1] \quad
    Wangze Xu\textsuperscript{1} \quad
    Qingtian Zhu\textsuperscript{2} \quad
    Muyao Niu\textsuperscript{2} \quad
    Mingze Ma\textsuperscript{2} \quad 
    Yifei Liu\textsuperscript{1,3} \quad \\[3pt]
    Zhihang Zhong\textsuperscript{1}\textsuperscript{†}\quad
    Xiao Sun\textsuperscript{1}\textsuperscript{†} \quad
    Yinqiang Zheng\textsuperscript{2} \\[8pt]
    \textsuperscript{1}Shanghai Artificial Intelligence Laboratory \qquad
    \textsuperscript{2}The University of Tokyo
    \qquad
    \textsuperscript{3}Beihang University
}

\begin{document}

\twocolumn[{
\renewcommand\twocolumn[1][]{#1}
\maketitle
\vspace{-3ex}
\centering
\thispagestyle{empty}
\includegraphics[width=1\linewidth\vspace{-1ex}]{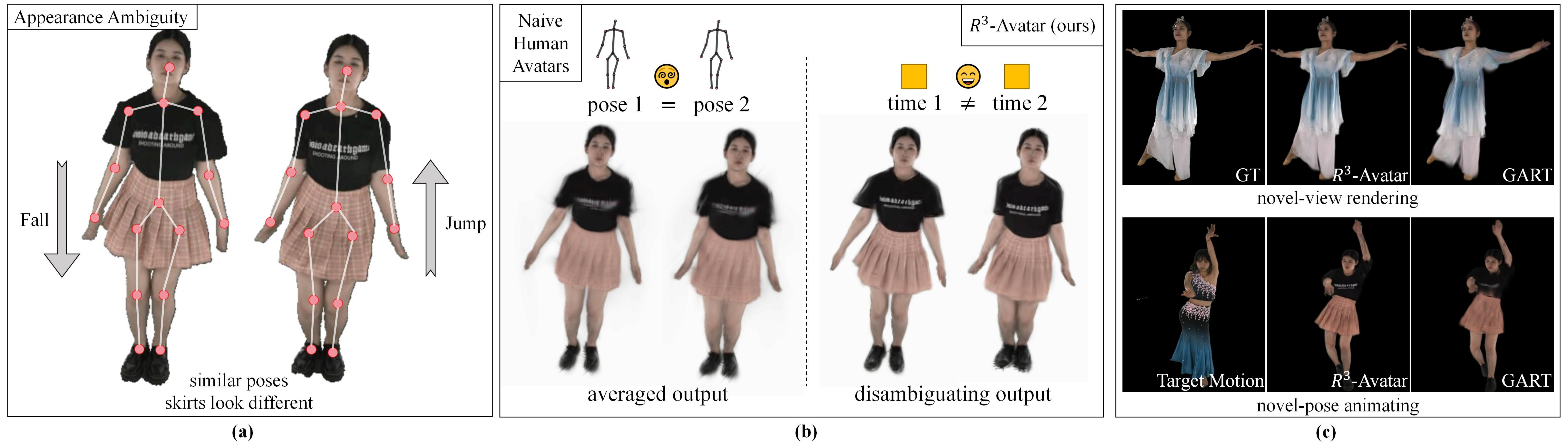}
\captionof{figure}{
(a) A case to show appearance ambiguity. The skirts of a same posed human look different due to different motion patterns (fall and jump) . (b) Our disambiguating ``record-retrieve-reconstruct'' strategy compared to the naive pose-input training pipeline, which is easily exposed to appearance ambiguity. (c) A glimpse of the results of the rendering and animation.}
\vspace{1ex}
\label{fig: teaser}
}]

\renewcommand{\thefootnote}{\fnsymbol{footnote}}
\footnotetext[1]{This work was done during the author's internship at the Shanghai Artificial Intelligence Laboratory. \textsuperscript{†} denotes co-corresponding authors.}

\begin{abstract}
We present $R^3$-Avatar, incorporating a temporal codebook, to overcome the inability of human avatars to be both animatable and of high-fidelity rendering quality.
% We propose \textbf{$\mathbf{R^3}$-Avatars}, a novel paradigm for reconstructing and animating the human gaussian avatar (HGS) from multi-view videos, designed to tackle challenges of fitting complex human clothing and generalizing to unseen human poses.
Existing video-based reconstruction of 3D human avatars either focuses solely on rendering, lacking animation support, or learns a pose-appearance mapping for animating, which degrades under limited training poses or complex clothing.
% Existing video-based reconstruction of 3D human avatars either focuses solely on rendering or relies on learning a pose-appearance mapping to generate novel appearance from unseen views and human poses.
% However, the former cannot support animation and downstream tasks, while the rendering and animation quality of later significantly degrades when the training set contains limited human poses or the clothing is excessively complex.
In this paper, we adopt a ``record-retrieve-reconstruct'' strategy that ensures high-quality rendering from novel views while mitigating degradation in novel poses.
Specifically, disambiguating timestamps record temporal appearance variations in a codebook, ensuring high-fidelity novel-view rendering, while novel poses retrieve corresponding timestamps by matching the most similar training poses for augmented appearance.
% During the recording phase, unambiguous timestamps serve as conditions to store the temporal variations in human appearance with a temporal codebook.
% In the retrieving phase, the novel poses serve as clues to search for the most similar training poses, allowing us to find the corresponding temporal feature for augmented appearance.
% Additionally, we design a pose-sequence-based coherent search algorithm that ensures the animated appearance remains sharp and temporally smooth.
Our $R^3$-Avatar outperforms cutting-edge video-based human avatar reconstruction, particularly in overcoming visual quality degradation in extreme scenarios with limited training human poses and complex clothing.
Code will be available at \href{https://github.com/Yifever20002/R3Avatars}{https://github.com/Yifever20002/R3Avatars}.
\end{abstract}

\thispagestyle{plain}

\section{Introduction}
\label{section: Introduction}

The emergence of 3D Gaussian Splatting~\cite{kerbl20233d} (3DGS) catches our attention once again to the reconstruction of digital human avatars.
Taking advantage of the rendering efficiency of 3DGS and its exceptional performance in novel view synthesis, recovering 3D avatars from human videos with gaussian representations~\cite{lei2024gart,li2024animatable,hu2024gauhuman,qian20243dgs,pang2024ash} outperforms earlier Neural Radiance Fields~\cite{mildenhall2020nerf} (NeRF)-based neural rendering methods~\cite{peng2023implicit,wang2022arah,weng2022humannerf,li2023posevocab}, raising expectations for enhanced rendering and animation quality in graphics and VR/AR technologies.

% However, these human gaussian avatar (HGS) methods still encounter challenges when addressing appearance ambiguities.
Existing modelings of digital human, however, struggle to escape the curse of being unable to balance both animating and rendering.
To achieve animation, human gaussians are constrained by pre-extracted human poses~\cite{hu2024gauhuman} or pose embeddings (\eg, a motion-aware map~\cite{pang2024ash}) to learn human appearance across different frames.
\cref{fig: teaser} (a) shows a typical one-to-many mapping, as noted by~\citet{chen2024within}, where similar poses are mapped to different appearances, thus limiting the modeling capacity of human gaussians.
Such ambiguities become more pronounced as the complexity of clothing increases.
Other rendering-based methods~\cite{lin2023im4d,xu20244k4d,jiang2024hifi4g,jiang2024robust,zheng2024gps,zhou2024gps} reconstruct per-frame human appearance without rigging a skeleton, limiting their ability to animate and perform other downstream tasks.

Since pose-appearance mapping is difficult to learn, could there be a ``mode-seeking'' approach that bypasses this learning process by first storing the appearance from the training set and then using the pose as a key to retrieving during animation?
% In this paper, we approach human gaussians modeling from a different perspective by leveraging disambiguating timestamps to learn variations of human appearance, rather than simply relying on human poses.
Accordingly, a ``record-retrieve-reconstruct'' strategy is implemented in this paper, where disambiguating recording of a temporal codebook enhances novel-view rendering, which is later retrieved by human poses to augment the appearance during animation.

For training during the recording phase, we maintain an explicit temporal codebook conditioned on timestamps to remember the variations of human appearance exploited in the provided video sequence.
Since human clothing exhibits non-rigid motions, we define the codebook in the T-pose space to decouple it from rigid motion, with the latter represented by the Linear Blend Skinning~\cite{loper2015smpl} (LBS) model.
Given the challenges of using a deformation field~\cite{qian20243dgs} to represent inter-frame motion of human clothing, we design a 4D gaussian decoder that directly outputs gaussian attributes in the canonical space for each timestamp.
In this way, the rendering at novel views could be guaranteed.

To tackle the lack of temporal information when performing animation under novel pose sequences, we propose a retrieving algorithm to determine the timestamps according to the similarity between novel poses and training poses.
However, directly matching the human poses of the whole body will fail due to limited matching samples.
% We first divide the human body into 5 parts based on DoF of human motion and match each part to its corresponding partial pose in the training set, allowing different body parts to have different timestamps.
To overcome this, we partition the human body, allowing different body parts to act asynchronously.
Then, we leverage pose sequences~\cite{chen2024within} to consider motion direction for more accurate matching.
To ensure visual coherence in animation, we further introduce a smoothing process.
The above components collectively form our retrieving algorithm.
% To ensure visual coherence for animating, it is desired the timestamps retrieved for novel poses to be continuous, just like those in the training set.
% However, this cannot be achieved through naive nearest-neighbor matching.
% We first aid this smoothing process by introducing human pose sequences to provide motion prior, thereby narrowing the matching space.
% Additionally, we set a searching window to limit abrupt changes in the timestamps corresponding to novel poses.
% At the same time, interpolation of pre-trained temporal feature is allowed to provide finer-grained appearance information.
% Finally, we introduce appearance smoothing to address the inevitable jumps of matched timestamps caused by out-of-distribution novel poses during the retrieval process.

We evaluate our method on multi-view datasets of humans with complex clothing, including DNA-Rendering dataset~\cite{cheng2023dna} and HiFi4G dataset~\cite{jiang2024hifi4g}.
To validate that our representation does not degrade in simpler clothing scenarios, we additionally compared it with the ZJU\_MoCap dataset~\cite{peng2021neural} and MVHumanNet dataset~\cite{xiong2024mvhumannet}.
To summarize, our contributions are three-fold as follows:
\begin{itemize}[noitemsep,topsep=0pt]
  \item [1)]
    a disambiguating training pipeline of human gaussians conditioned on timestamps, with a temporal codebook and a 4D gaussian decoder for efficient human appearance recording;
  \item [2)]
    a retrieving algorithm to obtain corresponding timestamps for novel poses, involving body-part decoupling, pose sequence matching, and appearance smoothing;
 \item [3)]
    state-of-the-art performance in both rendering and animating on datasets with diverse clothing complexities, compared to cutting-edge reconstructions of video-based 3D human avatars.
\end{itemize}

\section{Related Work}
\label{sec: Related_work}

\subsection{SMPL(-X) and Regression-based Human Avatars}
Most digital human works are built upon skinned vertex-based models such as SMPL(-X)~\cite{loper2015smpl, SMPL-X:2019}, which parameterizes the human body mesh as customized shape and motion-related joint poses.
The human body under different poses is constructed by linearly blending the mesh in the canonical space with the current joint poses.
This process is known as Linear Blend Skinning (LBS).
By the aid of the SMPL(-X) model and data-driven approaches, recovering the texture of a 3D human from images~\cite{saito2019pifu,he2020geo,saito2020pifuhd,huang2020arch,he2021arch++,xiu2022icon,xiu2023econ,zheng2021pamir} and videos~\cite{habermann2020deepcap,habermann2019livecap,xu2018monoperfcap,alldieck2018video,guo2021human} becomes possible.
However, SMPL is unable to model the motion of clothing, leading some works~\cite{casado2022pergamo,moon20223d} to focus on clothing reconstruction.

\subsection{Neural Animatable Human Avatars}
The emergence of NeRF~\cite{mildenhall2020nerf} has provided a new approach for 3D human avatar reconstruction.
Many NeRF-based methods backward-warp the human body to construct a neural radiance field of the human body in the canonical T-pose, leveraging human poses~\cite{peng2021neural,peng2024animatable,su2022danbo,wang2022arah,weng2022humannerf,su2021nerf,su2023npc,yu2023monohuman,li2022tava,li2023posevocab,zheng2022structured,jiang2022selfrecon,jiang2022neuman,guo2023vid2avatar,yin2023humanrecon,kim2024motion} or motion-aware maps~\cite{liu2021neural,instant_nvr,zhao2022high}.
Recent advancements in 3DGS~\cite{kerbl20233d} have further boosted the quality of neural rendering, leading to a surge in digital human research.
However, the rendering of clothing, as an important part of human avatars, has not been fully addressed.
Some human gaussian works~\cite{hu2024gauhuman,liu2024gva,paudel2024ihuman,shao2024splattingavatar,moon2024expressive,jung2023deformable,pang2024ash,kocabas2024hugs,jena2023splatarmor} ignore non-rigid appearance modeling while others~\cite{lei2024gart,li2024animatable,qian20243dgs,moon2024expressive} formulate the non-rigid part with human pose condition, not managing to escape from appearance ambiguities.
Generative methods have been introduced to enhance the rendering quality from unseen views~\cite{wang2025wonderhuman}.
We notice that frame index is used in~\cite{li2023human101} to alleviate such ambiguity trivially, incapable of accurately learning the variations in appearance and animating.
Pose sequence embedding proposed by~\cite{chen2024within, xu2024gast} serves as a disambiguating condition for appearance learning, while still suffering from the tricky selection of sequence length and steps, which is critical for training results.
% Our method, instead, utilizes the pose sequence solely during the animating phase to retrieve appearance, thereby exerting no influence on the training of the human avatar.
Our 4D gaussian decoder draws inspiration from~\cite{kocabas2024hugs,hu2024gaussianavatar,zou2024triplane}, and further incorporats a temporal codebook to acquire human gaussian attributes for each timestamp.

\subsection{Human Volume Rendering}
Rendering-based human avatars focus on recovering 3D human bodies from images or videos without the requirement for animation.
A class of methods~\cite{icsik2023humanrf,lin2023im4d,xu20244k4d,jiang2024hifi4g,jiang2024robust,zheng2024gps,zhou2024gps} employs multi-view image-based rendering strategies, using reference views to enhance the rendering quality from source views.
Other methods~\cite{ho2024sith,huang2024tech,albahar2023single,svitov2023dinar,alldieck2022photorealistic,zhang2024humanref,chen2024generalizable,pan2025humansplat,sun2024occfusion} aim to recover 3D textured humans from a single image with diffusion models.
\citet{svitov2023dinar} allow for simple animation, but limited to tight-fitting garments conforming to the SMPL geometry.
\citet{guo2024reloo} and \citet{tan2024dressrecon} can recover complex clothing from a monocular video without performing animation.
It is worth noting that the concept of a codebook~\cite{van2017neural, esser2021taming, zhou2022towards}, or look-up table~\cite{jo2021practical, li2020blind} (LUT), has been used to store information during training, enabling faster and more accurate inference.

% \subsection{Retrieval-Augmented Generation}
% While not directly related, we are inspired by a hint in RAG-related works~\cite{wang2024unims, luo2023reasoning, he2025g}, where external databases are incorporated to enhance LLMs overcoming hallucination and outdated knowledge.
% For our task, we considered whether external knowledge, such as fine-grained appearance from the training set, could enhance performance under novel poses.
% This led to the idea of using the training set as an appearance codebook for retrieval.

\section{Preliminaries}
\label{sec: Preliminaries}

\subsection{SMPL(-X)}
The Skinned Multi-Person Linear (SMPL)~\cite{loper2015smpl} model consists of human shape and pose parameters as well as per-vertex blending weights, which is learned from thousands of 3D human scans.
The canonical human mesh template $\bm{x}_c$ can be transformed into $\bm{x}_o$ in the observation space through LBS process, following
\begin{equation}
    \bm{x}_o = \sum_{k=1}^{K}\omega_k(\bm{x}_c)(R_k(\bm{\theta})\bm{x}_c + t_k(\bm{j}, \bm{\beta})),
\label{equ: LBS}
\end{equation}
where $K$ is the number of joints, $\omega_k$ is the per-joint linear blending weight, $R_k$ is the per-joint global rotation calculated by local human pose $\bm{\theta}$ and $t_k$ is the per-joint global translation calculated by joint position $\bm{j}$ and human shape $\bm{\beta}$.
Subsequent works extend the SMPL skeleton to include the hands~\cite{MANO:SIGGRAPHASIA:2017} and expressive faces~\cite{SMPL-X:2019}.
For human gaussian rendering, SMPL vertices provide a good gaussian initialization and blending weights. 
Nevertheless, clothed humans do not fully conform to the SMPL geometry, requiring the optimization of LBS weights.
Furthermore, recovering complex clothing through pose-based linear blending presents big challenges.

\subsection{3D Gaussian Splatting}
3DGS~\cite{kerbl20233d} represents 3D scenes by explicitly defining optimizable gaussian units.
Specifically, each gaussian unit can be defined as
\begin{equation}
\label{eqn:3d_gaussian_splatting}
\setlength{\abovedisplayskip}{5pt} 
\setlength{\belowdisplayskip}{5pt}
\begin{aligned}
G(\bm{x})=\frac{1}{(2\pi)^{\frac{3}{2}}|\bm{\Sigma}|^{\frac{1}{2}}}e^{-\frac{1}{2}(\bm{x}-\bm{\mu})^{T}\bm{\Sigma}^{-1}(\bm{x}-\bm{\mu})},
\end{aligned}
\end{equation}
where $\bm{\mu}$ is the gaussian center, and $\bm{\Sigma}$ is the 3D covariance matrix, which will be further decomposed into learnable rotation $\bm{R}$ and scale $\bm{S}$.
Other properties include gaussian color $c$ and opacity $\alpha$.

During the tile-based rendering, 3D gaussians are projected onto 2D, and the final screen color is determined based on $\alpha$-blending following
\begin{equation}
\setlength{\abovedisplayskip}{5pt} 
\setlength{\belowdisplayskip}{5pt}
\begin{aligned}
C=\sum_{i\in N} c_i\alpha_i\prod_{j=1}^{i-1}(1-\alpha_j).
\end{aligned}
\end{equation}
The gaussian properties will be optimized with image supervision.
An adaptive density control strategy is applied to the number of 3D gaussians through splitting and pruning. 
Furthermore, opacity is periodically reset to remove non-salient gaussians.

\subsection{Rigid Body Transformation}
\label{subsec: Rigid Body Transformation}

The canonical 3DGS attributes ($\bm{x}_c$, $\alpha$, $\bm{q}_c$, $\bm{s}$, $\text{sh}$) can be warped to the observation space using the LBS method.
Based on~\cref{equ: LBS}, the Euler transformation from the canonical space to the observation space is $T=\sum_{k=1}^{K}\omega_kT_k$, $T_k=\begin{bmatrix} R_k & t_k \\ 0 & 1 \end{bmatrix}$.
$T$ is applied to the gaussian geometric properties $\bm{x}_c$ and $\bm{q}_c$ according to
\begin{equation}
\bm{x}_o = T\bm{x}_c, \,\quad
\bm{R}_o = T\bm{R}_c,
\label{equ: 4d gaussian decoder}
\end{equation}
where $R$ is the matrix form of rotation $q$.

\section{Methods}
\label{sec: Methods}

\begin{figure*}[t]
\centering
\includegraphics[width=1\linewidth]{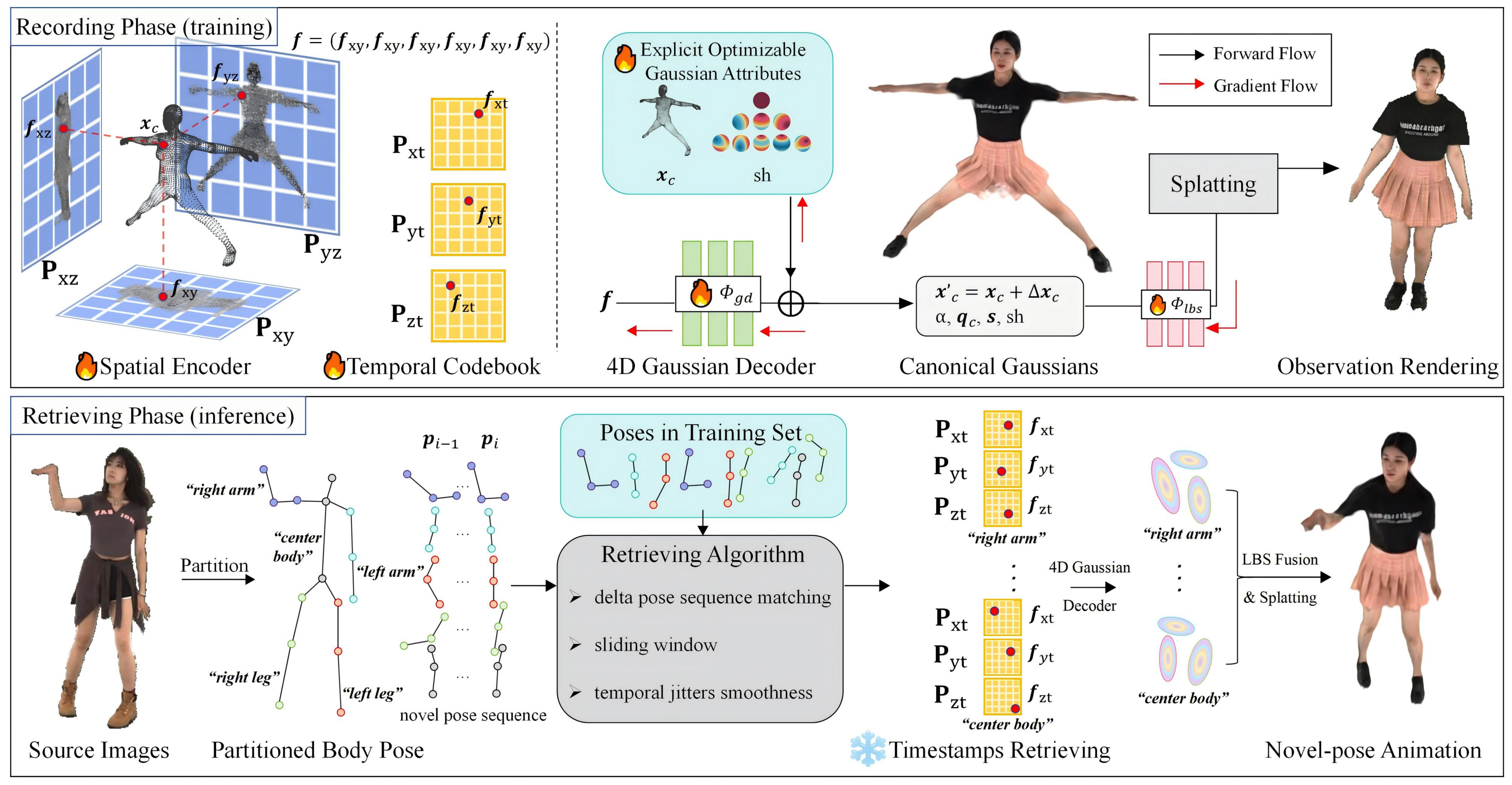}
    \vspace{-4ex}
    \caption{The pipeline of $R^3$-Avatar. In the ``recording phase'' (\cref{subsec: Multi-plane Spatio-temporal Encoder} and~\cref{subsec: 4D Gaussian Decoder}), a temporal codebook is used to capture human appearance variations over time. In the ``retrieving phase'' (\cref{subsec: Appearance Retrieving for Rendering}), retrieving the temporal codebook enables high-fidelity novel-view rendering and novel-pose animation.}
    \vspace{-2ex}
\label{fig: pipeline}
\end{figure*}

In this section, we will provide a detailed description of our ``record-retrieve-reconstruct'' strategy.
We first explain our training framework, where the time-related non-rigid appearance is \emph{recorded} by a hex-plane encoder and a 4D gaussian decoder, and the rigid spatial transformation is achieved through the LBS process.
Next, we will elaborate on our \emph{retrieving} strategy for both novel-view rendering and novel-pose animation, ensuring the accurate recovery of appearance while maintaining temporal smoothness during motion.
Please refer to \cref{fig: pipeline} for an overview of the entire pipeline.

\subsection{Hex-plane Spatio-temporal Encoder}
\label{subsec: Multi-plane Spatio-temporal Encoder}

We begin with a set of 3D points $\bm{x}_c\in\mathcal{R}^{3}$ defined in the canonical space and explore how to obtain the 3D gaussians in the observation space at each time frame $t\in\mathcal{R}^{1}$.
For an efficient encoding process, we adopt a hex-plane spatio-temporal encoder~\cite{fridovich2023k} with 6 orthogonal feature planes each in shape ${H}\times{W}\times{C}$, noted as ($\textbf{P}_{xy}$, $\textbf{P}_{xz}$, $\textbf{P}_{yz}$, $\textbf{P}_{xt}$, $\textbf{P}_{yt}$, $\textbf{P}_{zt}$).
Given a 4D input ($\bm{x}_c$, $t$), each feature plane is queried and bilinearly interpolated to obtain ($\bm{f}_{xy}$, $\bm{f}_{xz}$, $\bm{f}_{yz}$, $\bm{f}_{xt}$, $\bm{f}_{yt}$, $\bm{f}_{zt}$), which will be multiplied to get the spatio-temporal feature $\bm{f}\in\mathcal{R}^{C}$.
% Furthermore, we introduce multi-scale planes to effectively encode information at different frequencies, and concatenate them to obtain features at varying resolutions.
We expect the spatial planes ($\textbf{P}_{xy}$, $\textbf{P}_{xz}$, $\textbf{P}_{yz}$) to record the canonical static information, and the temporal planes ($\textbf{P}_{xt}$, $\textbf{P}_{yt}$, $\textbf{P}_{zt}$) to record the appearance variations caused by clothing at different time.

\subsection{4D Gaussian Decoder}
\label{subsec: 4D Gaussian Decoder}

The 4D gaussian decoder aims to recover 3D gaussians for each time in canonical space from spatio-temporal feature $f$.
We implement it through a simple yet efficient Multilayer Perceptron (MLP), whose input is $\bm{f}$ and the corresponding output is the canonical gaussian attributes for the current frame.
Our 4D gaussian decoder follows
\begin{equation}
    (\Delta\bm{x}_c,\alpha,\bm{q}_c,\bm{s})=\Phi_{gd}(\bm{f}).
\label{equ: 4d gaussian decoder}
\end{equation}
The canonical gaussian attributes include per-vertex offset $\Delta\bm{x}_c\in\mathcal{R}^{3}$, opacity $\alpha\in\mathcal{R}^{1}$, anisotropic covariance (decomposed into quaternion-represented rotation $\bm{q}_c\in\mathcal{R}^{4}$, and scale $\bm{s}\in\mathcal{R}^{3}$).

Note that some researches~\cite{zou2024triplane,hu2024gaussianavatar,wang2025wonderhuman,moon2024expressive,kocabas2024hugs} also use gaussian decoders to decode gaussian attributes from MLPs.
Here, we discuss the differences in the designs.
Non-rigid motions are not modeled in~\cite{kocabas2024hugs}, whose decoder assumes a static canonical space.
Although pose-dependent non-rigid deformation is considered in~\cite{zou2024triplane,hu2024gaussianavatar,wang2025wonderhuman,moon2024expressive}, the appearance ambiguity between human poses and clothing still hinders the decoder's learning capacity.
In contrast, our decoder adopts a human pose-independent 4D input, enabling disambiguating learning of appearance variations.

Another difference is that most gaussian decoders~\cite{zou2024triplane,hu2024gaussianavatar,wang2025wonderhuman,moon2024expressive} assume $\bm{q}_c=[1,0,0,0]$ and $\alpha=1$, and do not perform adaptive density control~\cite{kerbl20233d} for geometry regularization.
This works well for simple clothing, but experiments show that our design performs better when dealing with complex clothing.
Specifically, we allow $\bm{q}_c$ and $\alpha$ to be learned freely, and permit the adaptive pruning of $\bm{x}_c$.
Since color is not spatially continuous, we empirically find that learning spherical harmonics coefficients (sh) through per-gaussian optimization results better.
After obtaining the canonical 3DGS attributes ($\bm{x}'_c=\bm{x}_c+\Delta\bm{x}_c$, $\alpha$, $\bm{q}_c$, $\bm{s}$, $\text{sh}$), we follow~\cref{subsec: Rigid Body Transformation} to warp gaussians to the observation space.
The entire framework from~\cref{subsec: Multi-plane Spatio-temporal Encoder} to~\cref{subsec: 4D Gaussian Decoder} is optimized by the loss towards the rendered 3D gaussian attributes in the observation space ($\bm{x}_o$, $\alpha$, $\bm{q}_o$, $\bm{s}$, $\text{sh}$).
%For more details, please refer to~\cite{kerbl20233d}.

\begin{algorithm}[t]
\caption{Retrieving Algorithm}
\begin{algorithmic}

\Require \\
Novel delta pose $S^N_{\Delta{\bm{p}}}$,  \\
Training delta pose set $\mathcal{S} = \{S_{\Delta{\bm{p}_1}}, \dots, S_{\Delta{\bm{p}_t}}, \dots\}$.

\Ensure \\
To $S^N_{\Delta{\bm{p}}}$, the closest $S_{\Delta{\bm{p}_t}}$ in $\mathcal{S}$ and its timestamp $t$.

\vspace{2ex}
\State $d_i = \|S_{\Delta{\bm{p}_i}} - S^N_{\Delta{\bm{p}}}\|_2$,
\State $\mathcal{S}_{\text{top$k$}} = \{S_{\Delta{\bm{p}_{(1)}}}, \dots, S_{\Delta{\bm{p}_{(k)}}}\}, \quad d_{(1)} \leq \dots \leq d_{(k)}$.

\If{no historical retrieving $t_h$}
    \State $t \gets \operatorname{index}(\mathcal{S}_{\text{top}k}[0])$,
    \State $t_h \gets t$.
\Else
    \State Define sliding window size $W$,
    \State $\mathcal{S}_{\text{valid}}=\{S_{\Delta{\bm{p}_{(i)}}}\}$ where $|\operatorname{index}(\mathcal{S}_{\text{top}k}[i])- t_h| < W$.

    \If{$|\mathcal{S}_{\text{valid}}| > 2$}
        \State $t_1 \gets \operatorname{index}(\mathcal{S}_{\text{valid}}[0])$,
        \State $t_2 \gets \operatorname{index}(\mathcal{S}_{\text{valid}}[1])$,
        \State $t \gets \frac{d_{(2)}}{d_{(1)}+d_{(2)}}t_1+\frac{d_{(1)}}{d_{(1)}+d_{(2)}}t_2$,
        \State $t_h \gets t$.
        
    \ElsIf{$|\mathcal{S}_{\text{valid}}| = 1$}
        \State $t \gets \operatorname{index}(\mathcal{S}_{\text{valid}}[0])$,
        \State $t_h \gets t$.
    \Else
        \State $t \gets \operatorname{index}(\mathcal{S}_{\text{top}k}[0])$,
        \State $t_h \gets t$,
        \State Record a temporal jitter.
    \EndIf
\EndIf
\State \Return $t$
\end{algorithmic}
\label{alg:Appearance Retrieving Algorithm}
\end{algorithm}

\subsection{Appearance Retrieving for Rendering}
\label{subsec: Appearance Retrieving for Rendering}

The retrieving for novel-view rendering is simple as we already have the corresponding timestamps for novel-view images in place, thanks to multi-view synchronized capturing system.
In this section, we put up with a retrieving algorithm to handle animation for time-embedded human avatars.
Although our recording strategy can accurately capture the appearance in the training set, the animation ability under novel human poses are partly sacrificed as the timestamps for novel poses remain unknown.

A straightforward approach is to compare the similarity between the current novel pose $\bm{p}_i$ and all possible poses $\bm{p}^t$ in the training set $\bm{P}^t$, and to pick the timestamp of the frame with the most similar pose.
In practice, however, it is difficult to find a sufficiently close match of whole body pose (\eg, the matched pose may be similar overall, but with opposite arm movements).
Therefore, we first partition the body into 5 parts (namely center body $\bm{p}_{cb}$, left leg $\bm{p}_{ll}$, left arm $\bm{p}_{la}$, right leg $\bm{p}_{rl}$, right arm $\bm{p}_{ra}$), assuming that the movements among these parts could be relatively independent.
Now the matching samples are enriched and different parts can be retrieved with different timestamps.
In the following explanation, $\bm{p}_{i}$ will stand for per-part poses.

However, naive nearest-neighbor searching for each frame does not consider temporal smoothness, resulting in temporal jitters in the animation (\ie, the timestamps retrieved for adjacent novel poses are far apart in the training set).
To alleviate this, we first concatenate the $\bm{p}_{i}$ with delta pose sequence~\cite{chen2024within} to form $S_{\Delta{\bm{p}_{i}}}=\{\Delta{\bm{p}_{i-1}},\Delta{\bm{p}_{i}}, \bm{p}_{i}\}$ ($\Delta{\bm{p}_i} = \delta(\bm{p}_i, \bm{p}_{i-1})$) to ensure that the motion direction is also considered during matching.

% typical mode-seeking strategy
Furthermore, we set a sliding window to narrow down the candidates and to ensure that the next novel pose will prioritize timestamps near the time of the previous novel pose.
If the number of valid timestamps within the sliding window is greater than 1, we weight the 2 most similar timestamps by the respective similarity (inversely proportional to the L2 distance between poses) to obtain a more accurate retrieval time.
If there is only 1 valid timestamp within the sliding window, it will be selected directly.

In extreme cases where the input novel pose completely deviates from the training set distribution, we average the temporal features before and after the timestamps of these potential jitters to maintain the smoothness.
The entire retrieving algorithm can be referenced~\cref{alg:Appearance Retrieving Algorithm}.
We will thoroughly discuss the effectiveness of these modules and the ablation results in the supplemental materials.

\section{Experiments}
\label{sec: Experiments}

\subsection{Dataset}
\label{subsec: Dataset}

%novel view
\begin{table*}[ht]
\begingroup
\small
\begin{center}
\resizebox{\linewidth}{!}{
\begin{tabular}{lcccccccccccc}
\toprule
& \multicolumn{3}{c}{DNA-Rendering}             
& \multicolumn{3}{c}{HiFi4G}              
& \multicolumn{3}{c}{ZJU\_MoCap}
& \multicolumn{3}{c}{MVHumanNet} 
\\
\textbf{Method}   
& $\text{PSNR}\uparrow$          
& $\text{SSIM}\uparrow$          
& $\text{LPIPS}\downarrow$         
& $\text{PSNR}\uparrow$          
& $\text{SSIM}\uparrow$          
& $\text{LPIPS}\downarrow$
& $\text{PSNR}\uparrow$          
& $\text{SSIM}\uparrow$          
& $\text{LPIPS}\downarrow$         
& $\text{PSNR}\uparrow$          
& $\text{SSIM}\uparrow$          
& $\text{LPIPS}\downarrow$
\\ 
\cmidrule(r){1-1} 
\cmidrule(r){2-4} 
\cmidrule(r){5-7} 
\cmidrule(r){8-10}
\cmidrule(r){11-13}
4DGS~\cite{yang2023real}
& 20.27    & 0.900     & 87.8     
& 15.43    & 0.913     & 114.2  
& 22.98    & 0.874     & 131.0   
& 21.66    & 0.916     & 87.3
\\
Dyco~\cite{chen2024within}
& 28.06    & 0.959     & \cellcolor{orange!40}32.6     
& 27.36    & 0.968     & 30.2  
& \cellcolor{orange!40}33.77    & \cellcolor{orange!40}0.972     & 27.0   
& 31.47   &  0.973     & 19.2
\\
GART~\cite{lei2024gart}   
& \cellcolor{orange!40}29.62    & \cellcolor{orange!40}0.963     & 34.5      
& \cellcolor{orange!40}29.99    & \cellcolor{orange!40}0.973     & \cellcolor{orange!40}28.9  
& \cellcolor{yellow!40}33.61    & \cellcolor{orange!40}0.972     & \cellcolor{yellow!40}24.2  
& \cellcolor{orange!40}32.42    & \cellcolor{orange!40}0.975     & \cellcolor{yellow!40}18.8
\\
SplattingAvatar~\cite{shao2024splattingavatar}
& \cellcolor{yellow!40}28.83    & \cellcolor{yellow!40}0.961     & \cellcolor{yellow!40}34.0      
& \cellcolor{yellow!40}27.73    & \cellcolor{yellow!40}0.969     & \cellcolor{orange!40}28.9
& 33.06    & 0.971     & \cellcolor{red!40}21.3   
& \cellcolor{yellow!40}31.83    & \cellcolor{orange!40}0.975     & \cellcolor{orange!40}18.2
\\
$R^3$-Avatar (ours)
& \cellcolor{red!40}32.43    & \cellcolor{red!40}0.974     & \cellcolor{red!40}25.2 
& \cellcolor{red!40}32.18    & \cellcolor{red!40}0.983     & \cellcolor{red!40}20.7  
& \cellcolor{red!40}34.64    & \cellcolor{red!40}0.976     & \cellcolor{orange!40}22.5  
& \cellcolor{red!40}34.24    & \cellcolor{red!40}0.982     & \cellcolor{red!40}14.7
\\
\bottomrule
\end{tabular}
}
\end{center}
\endgroup
\vspace{-3ex}
\caption{Per-dataset quantitative comparisons on novel-view rendering. We color each result as \fcolorbox{red!40}{red!40}{best},  \fcolorbox{orange!40}{orange!40}{second best} and \fcolorbox{yellow!40}{yellow!40}{third best}. LPIPS is $1000\times$.}
\label{tab: Per-dataset quantitative comparisons}
\end{table*}

\begin{figure*}[h]
  \centering
  \includegraphics[width=1.\linewidth]{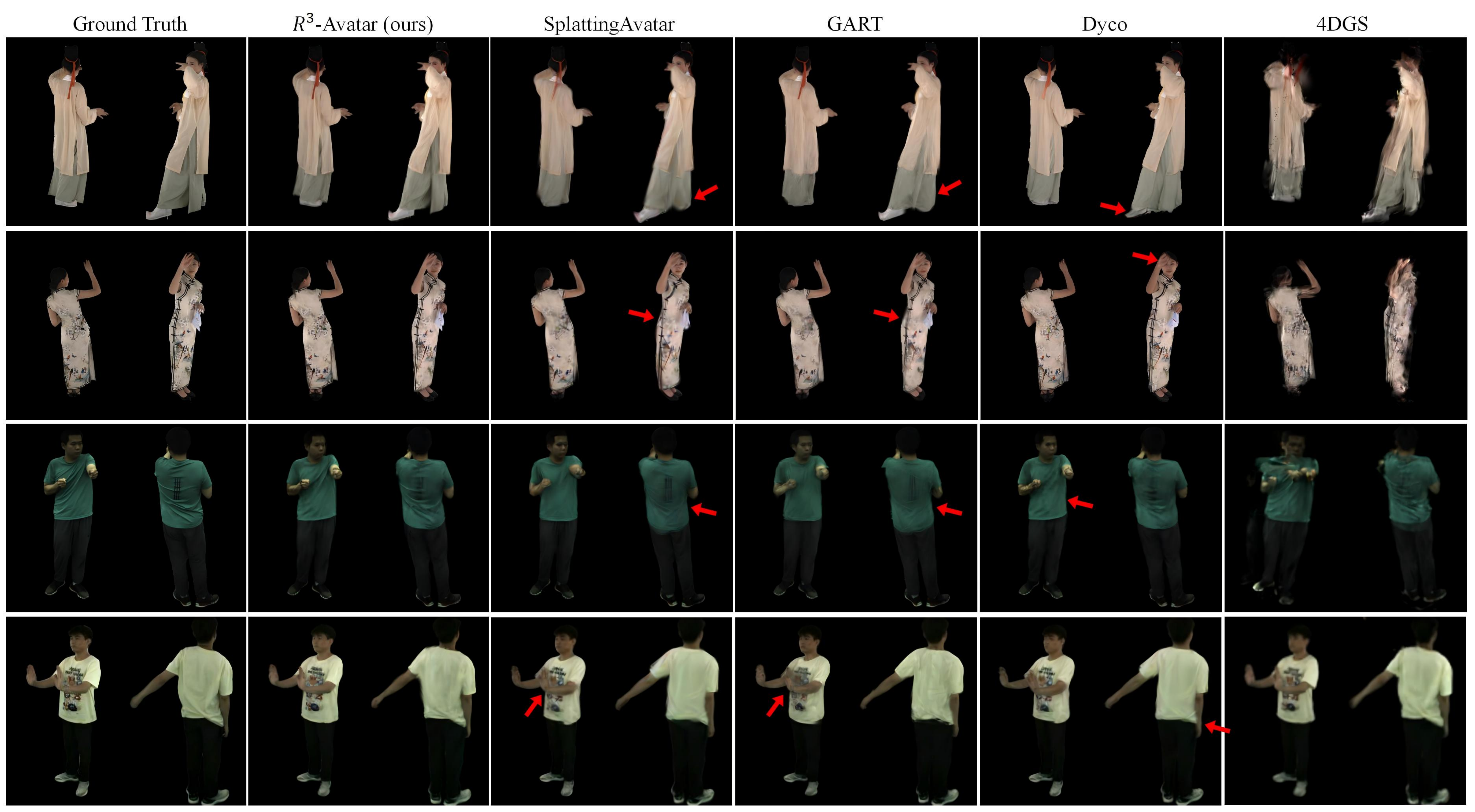}
      \vspace{-3ex}
      \caption{Novel-view rendering of our method and other baselines. 
      We compare on DNA-Rendering dataset~\cite{cheng2023dna} (first, second rows), ZJU\_MoCap dataset~\cite{peng2021neural} (third row) and MVHumanNet dataset~\cite{xiong2024mvhumannet} (last row) here.}
  \label{fig: novel-view rendering}
\end{figure*}

%novel pose
\begin{table*}[ht]
\begingroup
\small
\begin{center}
\resizebox{0.65\linewidth}{!}{
\begin{tabular}{lcccccccc}
\toprule
& \multicolumn{4}{c}{DNA-Rendering}            
& \multicolumn{4}{c}{MVHumanNet} 
\\
\textbf{Method}   
& $\text{PSNR}\uparrow$          
& $\text{SSIM}\uparrow$          
& $\text{LPIPS}\downarrow$         
& $\text{FID}\downarrow$    
& $\text{PSNR}\uparrow$          
& $\text{SSIM}\uparrow$          
& $\text{LPIPS}\downarrow$         
& $\text{FID}\downarrow$  
\\ 
\cmidrule(r){1-1} 
\cmidrule(r){2-5} 
\cmidrule(r){6-9} 
Dyco~\cite{chen2024within}
& \cellcolor{orange!40}25.00   & \cellcolor{orange!40}0.943   & \cellcolor{orange!40}48.5    & \cellcolor{orange!40}3.90 
& \cellcolor{orange!40}26.93   & \cellcolor{orange!40}0.952   & \cellcolor{yellow!40}34.5    & \cellcolor{orange!40}16.65
\\
GART~\cite{lei2024gart}   
& 24.26   & 0.941   & 49.6    & \cellcolor{yellow!40}4.25 
& 26.78   & \cellcolor{yellow!40}0.951   & 34.6    & 17.66
\\
SplattingAvatar~\cite{shao2024splattingavatar}
& \cellcolor{yellow!40}24.43   & \cellcolor{orange!40}0.943  & \cellcolor{yellow!40}49.4    & 4.28
& \cellcolor{yellow!40}26.87   & 0.950   & \cellcolor{orange!40}33.9    & \cellcolor{yellow!40}17.53
\\
$R^3$-Avatar (ours)
& \cellcolor{red!40}25.21   & \cellcolor{red!40}0.945   & \cellcolor{red!40}47.2    & \cellcolor{red!40}3.39
& \cellcolor{red!40}28.21   & \cellcolor{red!40}0.959   & \cellcolor{red!40}30.6    & \cellcolor{red!40}13.39 
\\
\bottomrule
\end{tabular}
}
\end{center}
\endgroup
\vspace{-3ex}
\caption{Per-dataset quantitative comparisons on novel-pose animating. We color each result as \fcolorbox{red!40}{red!40}{best},  \fcolorbox{orange!40}{orange!40}{second best} and \fcolorbox{yellow!40}{yellow!40}{third best}. LPIPS is $1000\times$.}
\label{tab: Per-dataset animating comparisons}
\end{table*}

\begin{figure*}[h]
  \centering
  \includegraphics[width=.9\linewidth]{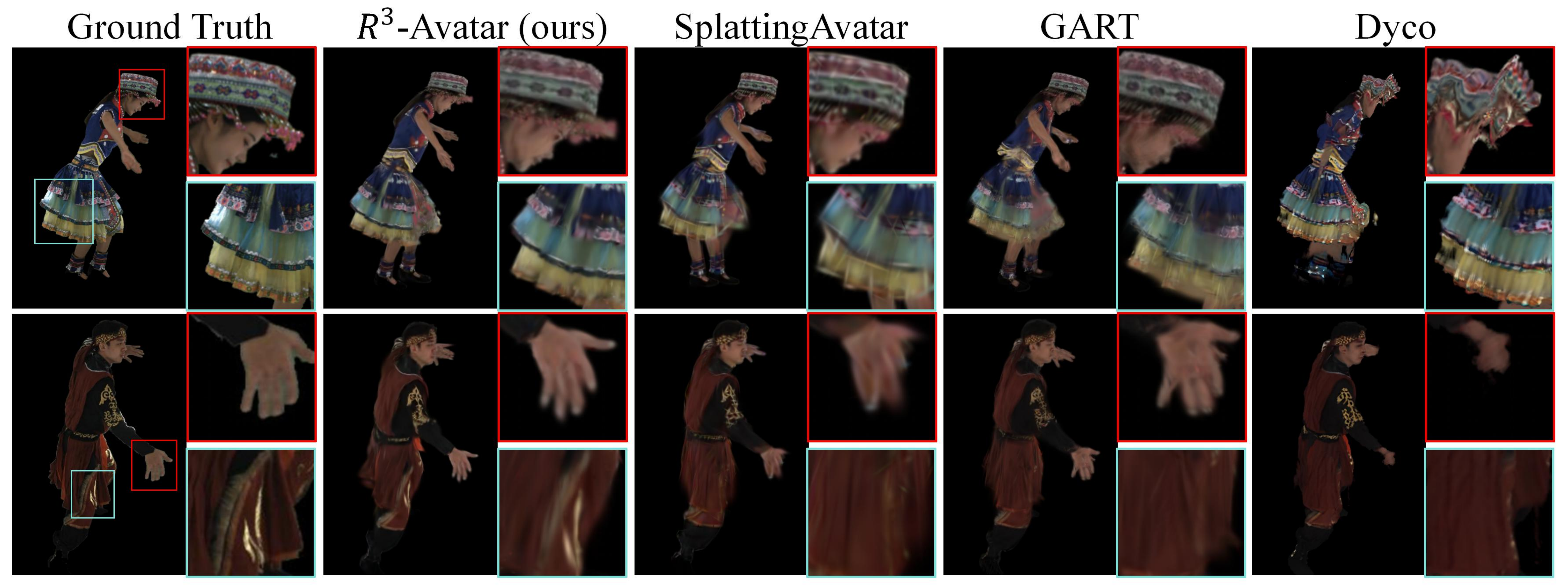}
      \vspace{-2ex}
      \caption{Novel-pose animating of our method and other baselines on DNA-Rendering dataset~\cite{cheng2023dna}.}
  \label{fig: novel-pose animating}
\vspace{-2ex}
\end{figure*}

\begin{figure}[h]
  \centering
  \includegraphics[width=1.\linewidth]{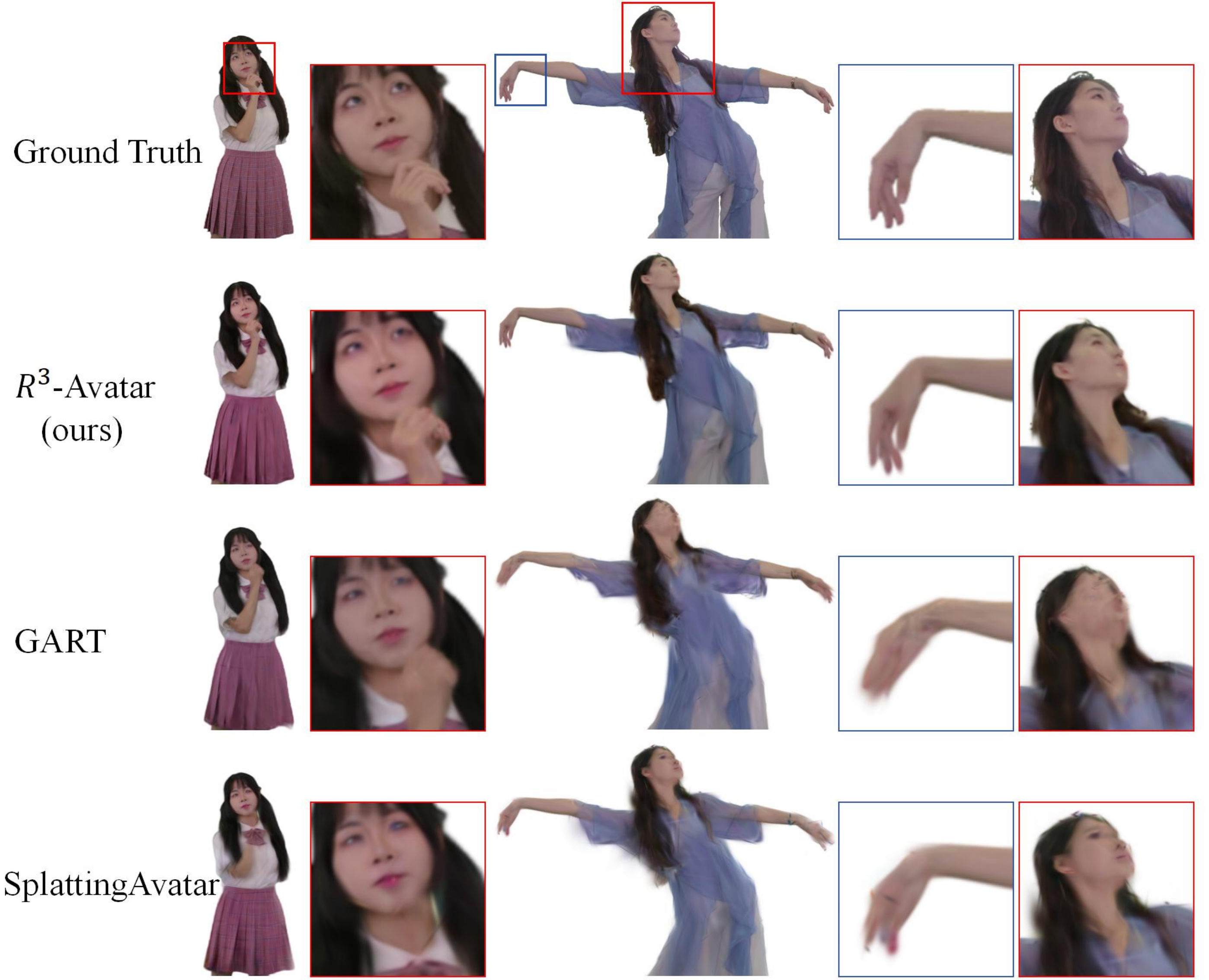}
      \caption{Novel-view rendering on HiFi4G dataset~\cite{jiang2024hifi4g}.}
  \label{fig: novel-view hifi}
\vspace{-2ex}
\end{figure}

Experiments are conducted on multi-view captured human datasets with complex clothing, namely DNA-Rendering dataset~\cite{cheng2023dna} and HiFi4G dataset~\cite{jiang2024hifi4g}.
For DNA-Rendering dataset, we select 24 surrounding views for training and 6 novel surrounding views for testing.
7 sequences (\emph{0010\_03, 0012\_03, 0019\_10, 0024\_05, 0051\_09, 0172\_05, 0188\_02}) are considered, each containing frame [0,100) for training frames and [110,150) for novel human poses.
For HiFi4G dataset, we select 17 surrounding views for training and 11 novel surrounding views for testing.
2 sequences (\emph{4K\_Actor1\_Greeting, 4K\_Actor2\_Dancing}) are considered, each containing frame [0,140) for training frames and [150,200) for novel human poses.

We also work on two multi-view human datasets with simpler clothing, namely ZJU\_MoCap dataset~\cite{peng2021neural} and MVHumanNet dataset~\cite{xiong2024mvhumannet} to see whether the proposed method will cause degradation.
For ZJU\_MoCap dataset, we select 8 surrounding views for training and 15 novel surrounding views for testing.
6 sequences (\emph{377, 386, 387, 392, 393, 394}) are considered, each containing frame [0,300) for training frames and [300,500) for novel human poses.
For MVHumanNet dataset dataset, we select 22 surrounding views for training and 9 novel surrounding views for testing.
4 sequences (\emph{100003, 100009, 100022, 100025}) are considered, each containing frame [0,150) for training frames (take one frame every five frames) and [150,300) for novel human poses.

\subsection{Baselines and Metrics}
\label{subsec: Baselines and Metrics}

We compare with 4DGS~\cite{yang2023real} to validate that incorporating human modeling is meaningful for reconstruction of human avatars under sparse views.
For human gaussian methods, we compare most cutting-edge methods including GART~\cite{lei2024gart} and SplattingAvatar~\cite{shao2024splattingavatar}.
We also compare pose sequence embedding-based method~\cite{chen2024within} to show that our ``record-retrieve-reconstruct'' approach is better in dealing with appearance ambiguity.
Yet not fully within the scope of our paper, we will also compare our rendering quality with non-animatable human rendering methods~\cite{lin2023im4d,zhou2024gps} and some methods~\cite{tan2024dressrecon,wang2025wonderhuman} for recovering unseen views by incorporating priors in monocular setups, in the supplemental materials.

For novel-view rendering, we report three key metrics: peak signal-to-noise ratio (PSNR), structural similarity index measure (SSIM)~\cite{wang2004image}, and learned perceptual image patch similarity (LPIPS)~\cite{zhang2018unreasonable}.
For novel-pose animating, we also report Frechet Inception Distance score (FID)~\cite{heusel2017gans}.
We encourage the readers to watch the supplemental video to gain a more intuitive understanding of the animation.

\subsection{Comparison of Rendering and Animating}
\label{subsec: Novel-View Rendering}

\noindent\textbf{Novel-view rendering}.
In~\cref{fig: novel-view rendering},~\cref{fig: novel-view hifi} and~\cref{tab: Per-dataset quantitative comparisons} we display results on novel-view rendering of our methods and compared baselines.
Our method achieves high-fidelity rendering results in both complex garments (first, second rows) and tight-fitting clothing (third, last rows) cases.
Please refer to the supplemental video for better visualization.

\noindent\textbf{Novel-pose animating}.
In~\cref{fig: novel-pose animating} and~\cref{tab: Per-dataset animating comparisons} we show results on novel-pose animating of our methods and compared baselines.
Our method maintains strong animating performance, especially when handling complex geometry and textures.
To clarify, novel poses presented here belong to ``in-distribution'' novel poses with ground truths, as they are clipped from the same video of the training avatar (see~\cref{subsec: Dataset} for detail split).

\subsection{Ablation Studies}
\label{subsec: Ablation Studies}

\begin{figure*}[h]
  \centering
  \includegraphics[width=1.\linewidth]{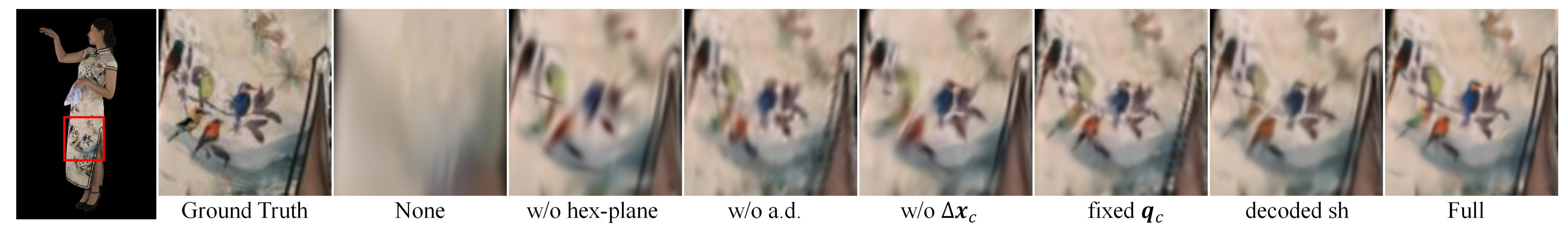}
      \vspace{-4ex}
      \caption{Ablations on the architecture of recording phase.}
  \label{fig: ablation_recording}
\end{figure*}

\noindent\textbf{The architecture of recording phase}.
We design an efficient training pipeline for recording temporally varying appearance.
From~\cref{tab: ablation of the recording phase architecture} and~\cref{fig: ablation_recording}, we observe a decline in metrics when ablating individual components.
Specifically, we test by replacing hex-plane encoder with positional encoding~\cite{mildenhall2020nerf, tancik2020fourier} P.E.$(\bm{x}, t)$ (w/o hex-plane), using 4d gaussian decoder to decode spherical harmonics coefficients (decoded sh), fixing rotations (fixed $\bm{q}_c$), not decoding $\Delta{\bm{x}}$ (w/o $\Delta{\bm{x}}$), not applying adaptive density control~\cite{kerbl20233d} (w/o a.d.) and implementing above all (None).
Among them, (w/o hex-plane) results in the most significant degradation and (decoded sh, fixed $\bm{q}_c$, w/o $\Delta{\bm{x}}$, w/o a.d.) lead to varying degrees of losing detail and blurring.
Finally, the complete removing of these components makes the human avatar entirely inaccessible.

\noindent\textbf{Design of retrieving algorithm}.
Our retrieving algorithm is designed to provide high-fidelity and smooth animation.
To ablate, we consider not partitioning human body (w/o b.p.),  not introducing sliding window (w/o s.w.), not weighting valid timestamps by their similarity (w/o w.t.) and ignoring temporal jitters smoothness (w/o t.s.).

\begin{table}[t]
\small
\begin{center}
\resizebox{0.7\linewidth}{!}{
\begin{tabular}{lccc}
\toprule
Component 
& $\text{PSNR}\uparrow$   
& $\text{SSIM}\uparrow$    
& $\text{LPIPS}\downarrow$\\
\cmidrule(r){1-1} 
\cmidrule(r){2-4}
None
& 29.13
& 0.958
& 53.0 \\
w/o a.d.
& 32.05
& 0.972
& 30.1 \\
w/o $\Delta{\bm{x}}$
& 31.91
& 0.971
& 31.9 \\
fixed $\bm{q}_c$
& 32.27
& 0.973
& 28.5 \\
decoded sh
& 32.06
& 0.973
& 27.0 \\
w/o hex-plane
& 30.46
& 0.966
& 36.8 \\
Full
& \textbf{32.43}
& \textbf{0.974}
& \textbf{25.2} \\
\bottomrule
\end{tabular}}
\end{center}
\vspace{-4ex}
\caption{
    \label{tab: ablation of the recording phase architecture}{Ablations on the architecture of recording phase on DNA-Rendering dataset~\cite{cheng2023dna}. LPIPS is $1000\times$.}
\vspace{-3ex}
}

\end{table}

\begin{figure*}[h]
  \centering
  \includegraphics[width=1.\linewidth]{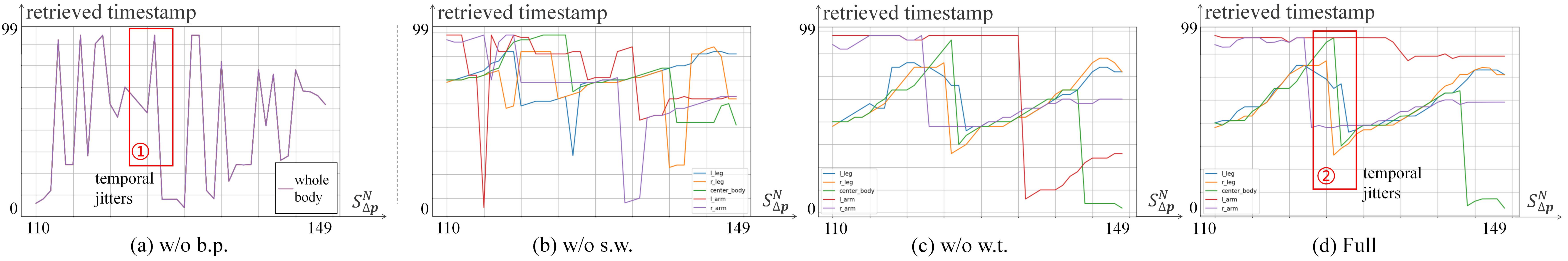}
      \vspace{-4ex}
      \caption{Ablations on design of retrieving algorithm. We show (retrieved timestamps)-(delta pose sequence) curve.}
  \label{fig: ablation_retrieving}
\vspace{-2ex}
\end{figure*}

In~\cref{fig: ablation_retrieving}, we take         ``dna\_0051\_09'' as an example to illustrate the impact of different designs on overall retrieving. 
The horizontal axis represents the index of the delta pose sequence $S^N_{\Delta{\bm{p}}}$ ranging from 110 to 150, while the vertical axis denotes the retrieved timestamps in training delta pose set $\mathcal{S}$.
In (a), as there is no body parts, the whole body joints follow inaccurate identical retrieved timestamps, which also results in drastic temporal jitters \textcolor{red}{\ding{172}}.
In (b), although we allow free retrieving for different body parts, the absence of a sliding window to optimize continuity of the novel appearance still leads to frequent temporal jitters.
In (c), since the retrieved timestamp can only be selected from discrete training timestamps, it fails to leverage the interpolability of the temporal codebook, compromising the smoothness of the output.
In (d), although smoothness is ensured, the occurrence of out-of-distribution poses still gives rise to temporal jitters, such as \textcolor{red}{\ding{173}}.

\begin{figure}[h]
  \centering
  \includegraphics[width=1.\linewidth]{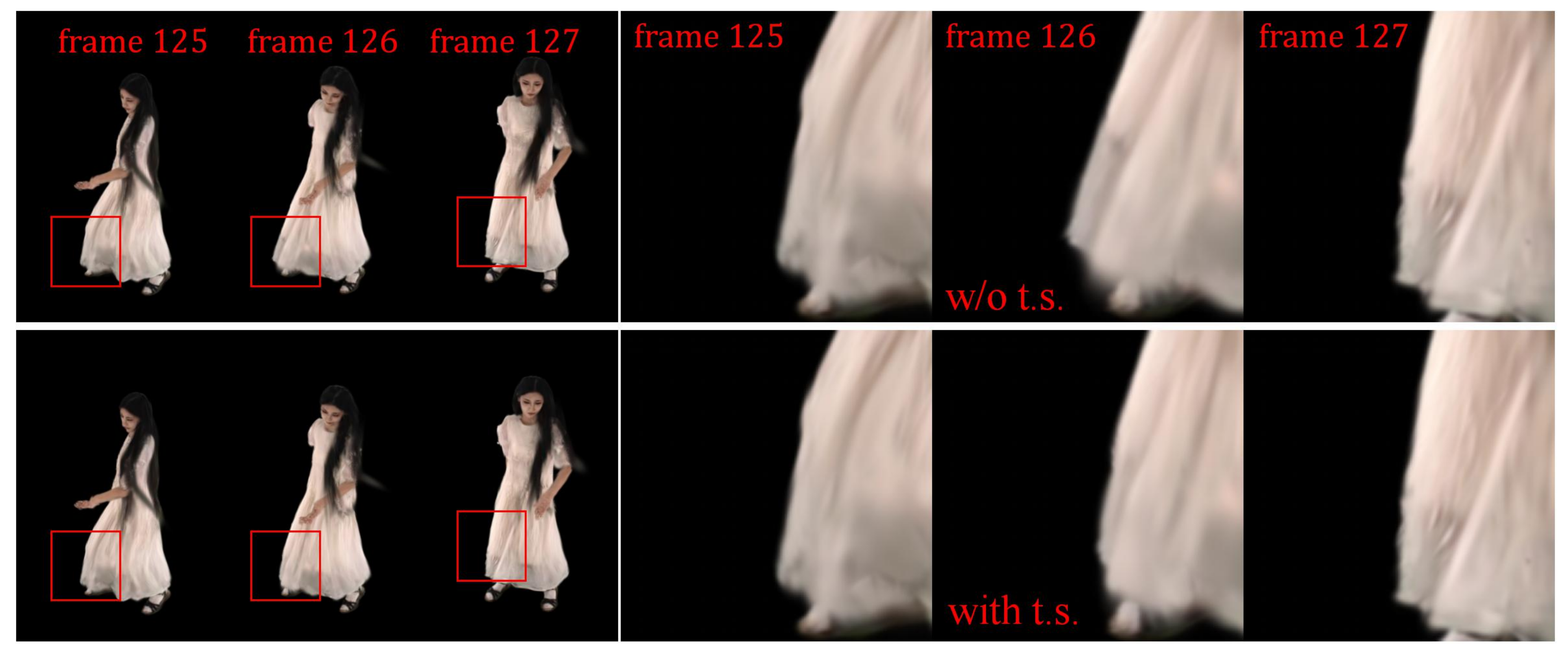}
      \vspace{-4ex}
      \caption{Without smoothing temporal jitters (the first row), the appearance exhibits a drastic change at frame 126, whereas the smoothness of temporal jitters ensures a more seamless animation of the driven avatar (second row).}
  \label{fig: ts}
\vspace{-2ex}
\end{figure}

In~\cref{fig: ts}, we demonstrate the effectiveness of temporal jitters smoothness.
If there is no smoothness (row 1) at frame 126 where temporal jitters happen, the appearance undergoes drastic changes , whereas with temporal smoothness (row 2) , it transitions smoothly.

\subsection{Out-of-distribution Animation}
\label{subsec: Out-of-distribution Animation}

\begin{figure}[t]
  \centering
  \includegraphics[width=1.\linewidth]{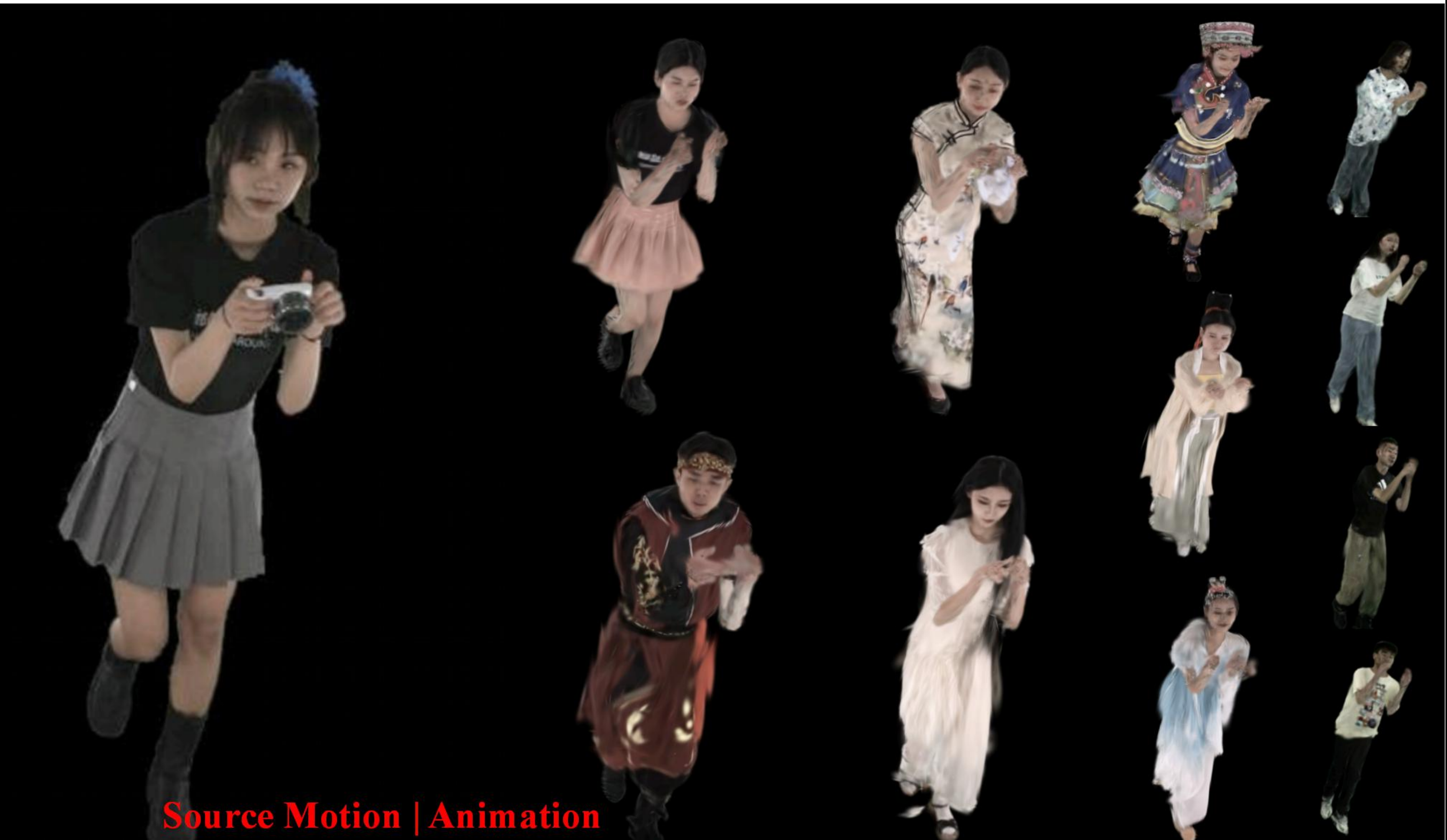}
      \vspace{-4ex}
      \caption{Animation of human assets under extreme out-of-distribution novel poses.}
  \label{fig: out-of-distribution}
\vspace{-3ex}
\end{figure}

In~\cref{fig: out-of-distribution}, we further demonstrate the animation under completely out-of-distribution poses.
Since no similar motions were observed during training, this animation is particularly challenging.

\section{Limitations and Conclusion}
\label{sec: Limitations and Conclusion}

\textbf{Limitations}.
Although our method can recover high-fidelity human appearance, it relies on multi-view video as inputs.
This imposes higher requirements on the capturing equipment, necessitating multiple synchronized cameras covering various viewpoints.
Research on recovering human avatars from monocular videos could be a future work, which remains highly challenging.
Many works~\cite{hu2024gauhuman, qian20243dgs, wang2025wonderhuman} claim to achieve monocular recovery, but they often require videos where the actor fully exposes all body parts (both front and back) or rely on diffusion-based completion to handle unseen views.
In the supplemental materials, we include a comparison with monocular methods to validate two key points.
First, existing monocular methods degrade significantly on challenging monocular videos. 
Second, our method still achieves comparable performance on simpler monocular videos.

\textbf{Conclusion}.
In this paper, we introduce $R^3$-Avatar, a temporal codebook-based framework that effectively addresses the long-standing challenge of balancing high-fidelity rendering and animatability in human avatar reconstruction.
By adopting a ``record-retrieve-reconstruct'' strategy, our method ensures high-quality novel-view rendering while mitigating degradation in novel-pose animation.
Specifically, we leverage disambiguating timestamps to encode temporal appearance variations in a codebook and retrieve the most relevant timestamps based on body-part-level pose similarity, enabling robust generalization to unseen poses.
Extensive experiments demonstrate that $R^3$-Avatar outperforms state-of-the-art video-based human avatar reconstruction approaches, particularly in extreme scenarios with few training poses and complex clothing.
Our work aims to inspire advancements in human avatar reconstruction, encouraging further exploration into complex clothing representations, which holds significant potential for broader applications in immersive 3D environments such as AR and VR.

{
    \small
    \bibliographystyle{ieeenat_fullname}
    \bibliography{main}
}

\newpage
\appendix
\section{Supplemental Material}

\subsection{Comparison under Monocular Setting}
\label{subsec: Comparison on Monocular Setting}
Our method exhibits degraded performance on monocular videos.
The main reason is that temporal encoding allows different frames to have varying appearances.
When multi-view constraints are insufficient, the degrees of freedom for unseen views increase.
However, we point out that some methods~\cite{hu2024gauhuman, tan2024dressrecon} claiming to recover the human body from monocular videos rely on highly constrained input, requiring the person to fully expose their body parts (both front and back) to the camera.

\begin{figure}[h]
  \centering
  \includegraphics[width=1.\linewidth]{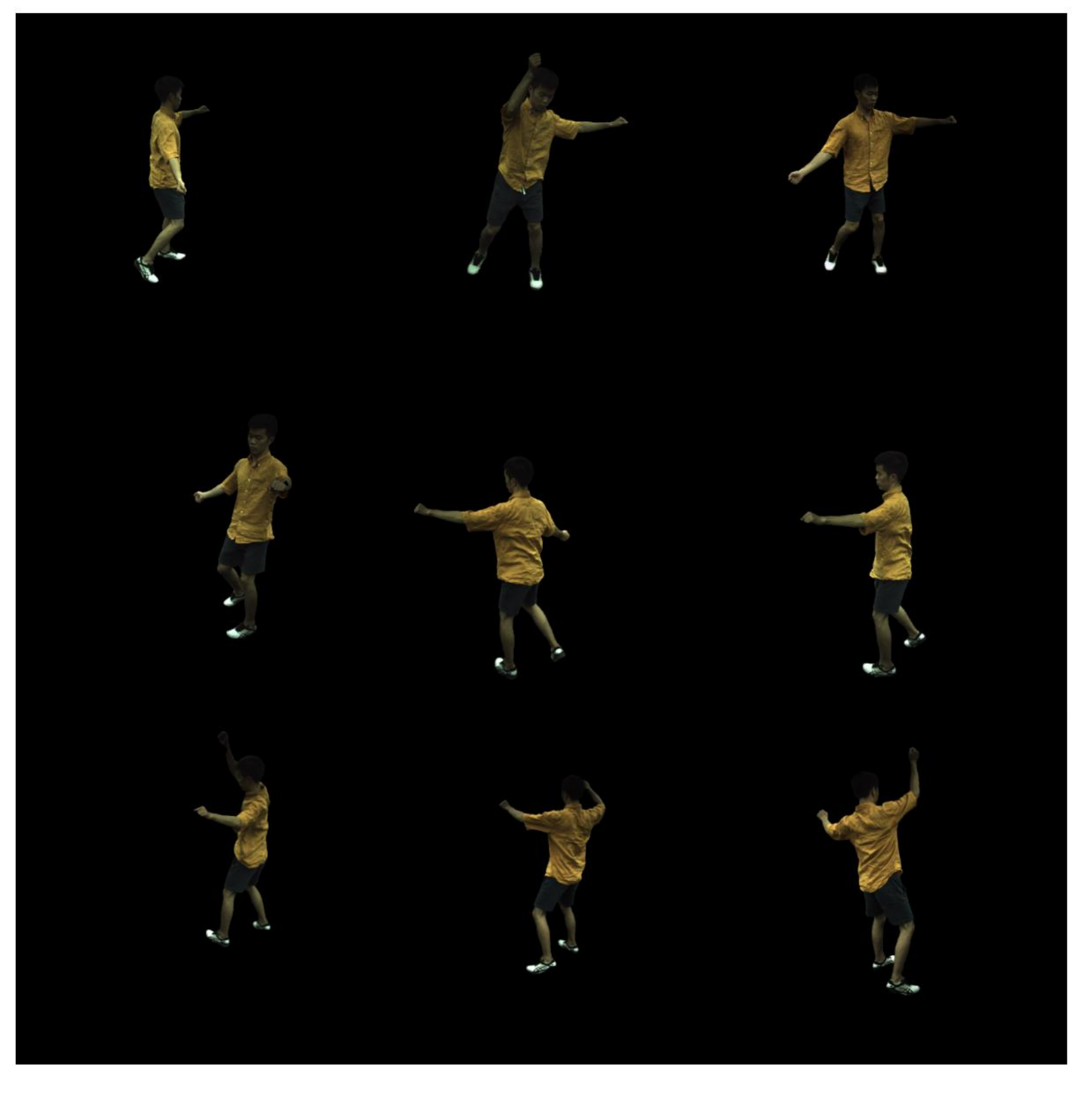}
  \vspace{-4ex}
  \caption{A monocular video example from the highly constrained ZJU\_MoCap dataset. The person is fully exposed to the camera.}
  \label{fig: zju}
\end{figure}

\begin{figure}[t]
  \centering
  \includegraphics[width=1.\linewidth]{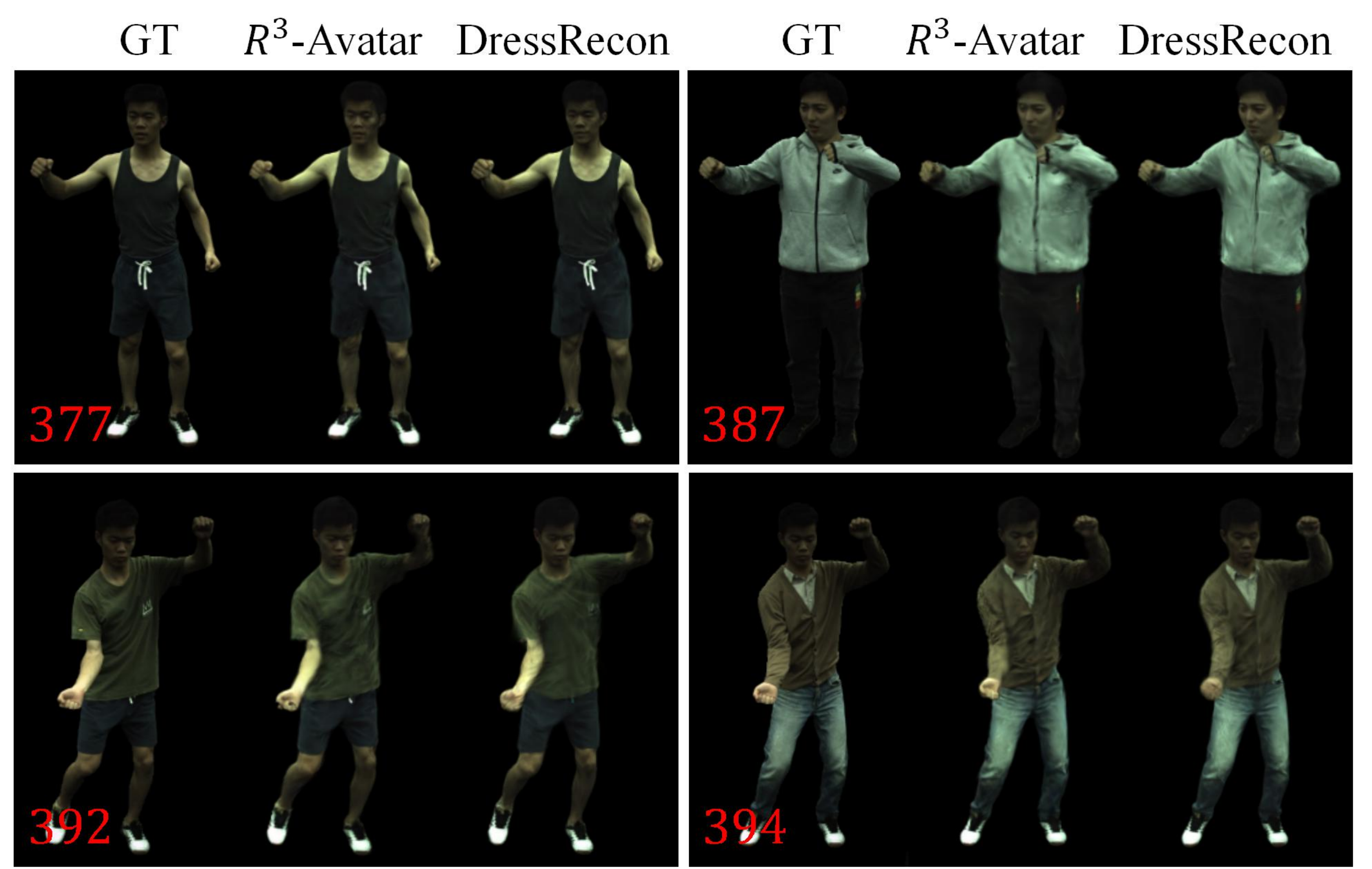}
  \vspace{-3ex}
  \caption{Qualitative comparison of our method and DressRecon on the ZJU\_MoCap dataset~\cite{peng2021neural}.}
  \label{fig: com_zju}
\end{figure}

In~\cref{fig: zju}, we show an example of ZJU\_MoCap dataset~\cite{peng2021neural} to illustrate this strict capturing.
And in the cases shown in~\cref{tab:per-scene quantitative on ZJU} and~\cref{fig: com_zju} (ZJU\_MoCap dataset trained from monocular video), our method performs slightly worse than the monocular-based DressRecon~\cite{tan2024dressrecon}, whichlacks animatability.

\begin{table*}[ht]
\begingroup
\small
\begin{center}
\resizebox{0.8\linewidth}{!}{
\begin{tabular}{lccccccccc}
\toprule
& \multicolumn{3}{c}{377}              & \multicolumn{3}{c}{386}              & \multicolumn{3}{c}{387}              \\
\textbf{Method}   
& $\text{PSNR}\uparrow$          
& $\text{SSIM}\uparrow$          
& $\text{LPIPS}\downarrow$         
& $\text{PSNR}\uparrow$          
& $\text{SSIM}\uparrow$          
& $\text{LPIPS}\downarrow$         
& $\text{PSNR}\uparrow$          
& $\text{SSIM}\uparrow$          
& $\text{LPIPS}\downarrow$         
\\ 
\cmidrule(r){1-1} \cmidrule(r){2-4} \cmidrule(r){5-7} \cmidrule(r){8-10}
DressRecon~\cite{tan2024dressrecon}   
& \textbf{31.30}    & \textbf{0.971}    & \textbf{21.2}      & 31.96  & 0.963    & \textbf{34.2}    & 27.60      & \textbf{0.952}
& \textbf{38.9}
\\
$R^3$-Avatar (ours)
& 30.82    & 0.968    & 30.0      & \textbf{32.41}  & \textbf{0.964}    & 40.0    & \textbf{27.79}      & 0.951
& 46.8
\\
\hline
& \multicolumn{3}{c}{392}              & \multicolumn{3}{c}{393}              & \multicolumn{3}{c}{394}              \\
\textbf{Method}
& $\text{PSNR}\uparrow$          
& $\text{SSIM}\uparrow$          
& $\text{LPIPS}\downarrow$         
& $\text{PSNR}\uparrow$          
& $\text{SSIM}\uparrow$          
& $\text{LPIPS}\downarrow$         
& $\text{PSNR}\uparrow$          
& $\text{SSIM}\uparrow$          
& $\text{LPIPS}\downarrow$         
\\ \cmidrule(r){1-1} \cmidrule(r){2-4} \cmidrule(r){5-7} \cmidrule(r){8-10}
DressRecon~\cite{tan2024dressrecon} 
& \textbf{31.94}    & \textbf{0.964}    & \textbf{31.0}      & \textbf{29.81}  & \textbf{0.956}    & \textbf{34.6}    & 31.02      & \textbf{0.958}
& \textbf{30.2}
\\
$R^3$-Avatar (ours)
& 31.92    & 0.961    & 39.8      & 29.40  & 0.951    & 44.3    & \textbf{31.06}      & 0.956
& 39.2
\\
\bottomrule
\end{tabular}
}
\end{center}
\endgroup
\vspace{-3ex}
\caption{Per-scene quantitative comparisons on the monocular ZJU\_MoCap dataset~\cite{peng2021neural}.}
\label{tab:per-scene quantitative on ZJU}
\end{table*}

\begin{figure}[t]
  \centering
  \includegraphics[width=1.\linewidth]{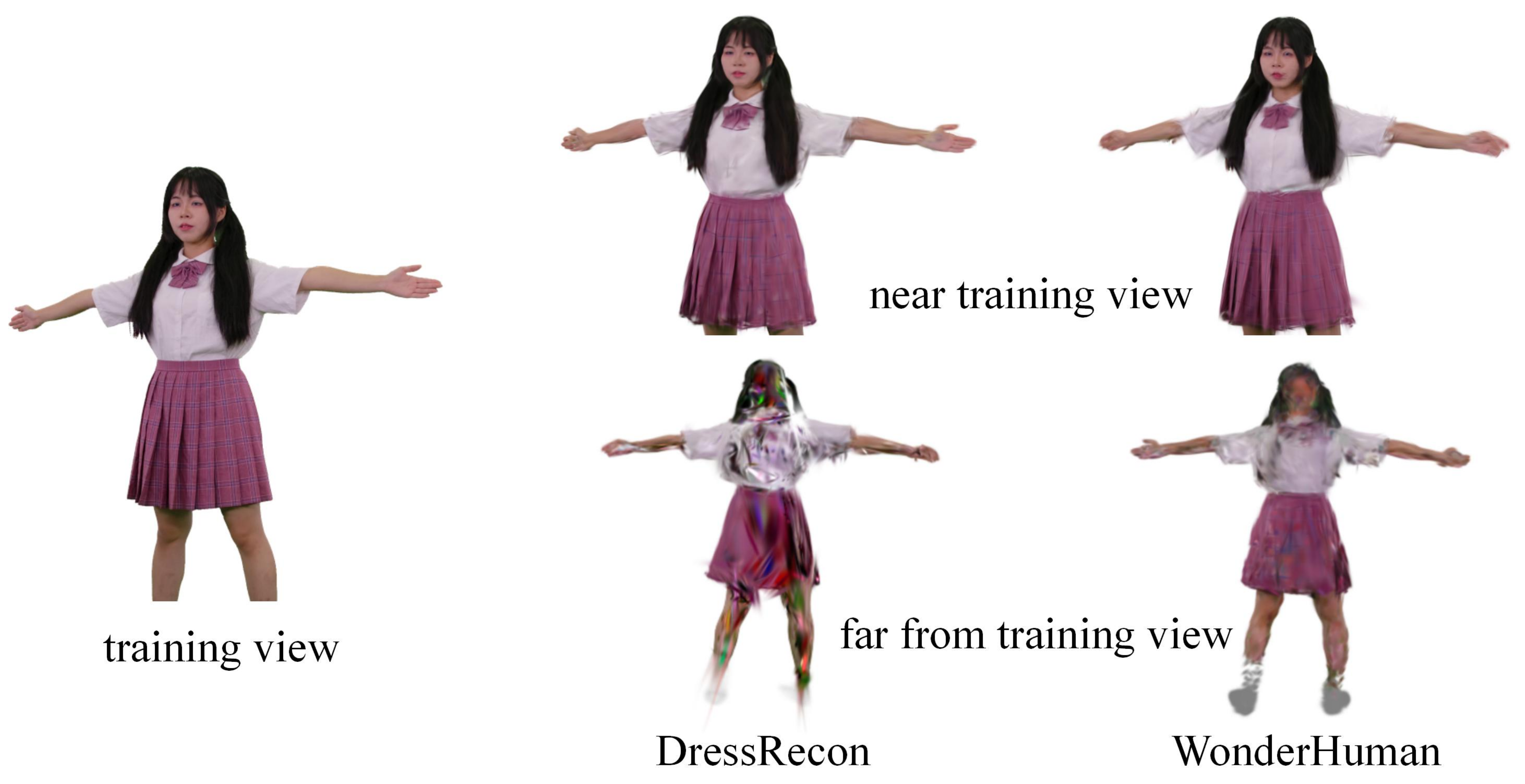}
  \caption{Qualitative results of DressRecon and WonderHuman on unconstrained capture data.}
  \label{fig: com_hifi}
\vspace{-3ex}
\end{figure}

\begin{figure*}[t]
  \centering
  \includegraphics[width=1.\linewidth]{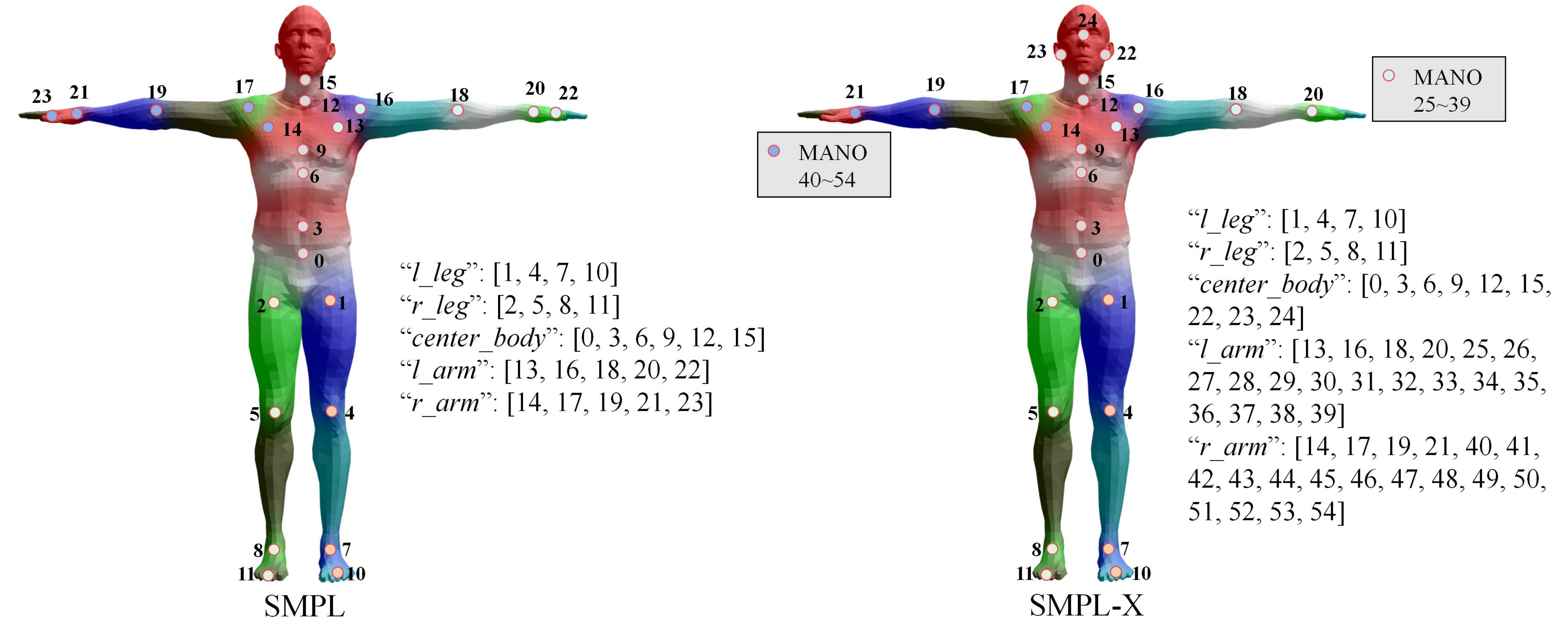}
  \caption{The body partitioning during retrieving on both SMPL (ZJU\_MoCap dataset, Hifi4G dataset) and SMPL-X (DNA-Rendering dataset, MVHumanNet dataset).}
  \label{fig: body_seg}
  \vspace{-2ex}
\end{figure*}

We further apply monocular methods for reconstruction on unconstrained capture data. 
As shown in~\cref{fig: com_hifi}, the reconstruction quality significantly deteriorates, indicating that these methods are not capable of perfectly handling all monocular data.
Our multi-view setting circumvents these challenges, allowing us to focus on human rendering and animation.
That said, recovering human bodies from imperfectly captured monocular videos remains a challenge.

\subsection{Comparison with Rendering-based Methods}
\label{subsec: Comparison on Rendering-based Methods}

\begin{figure}[t]
  \centering
  \includegraphics[width=1.\linewidth]{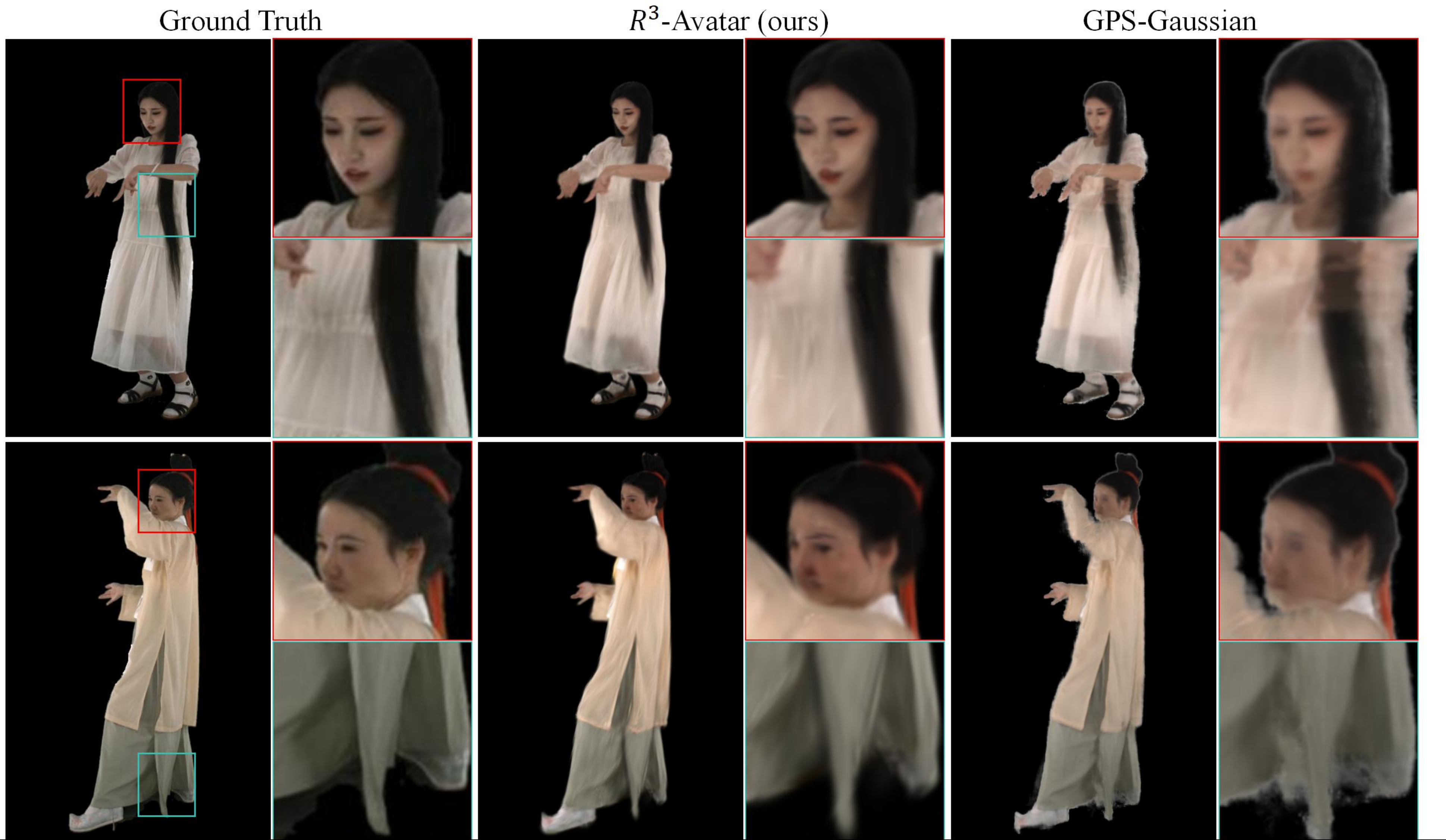}
  \caption{Novel-view rendering comparison of our method and GPS-Gaussian on DNA-Rendering dataset~\cite{cheng2023dna}.}
  \label{fig: gps}
\end{figure}

Rendering-based human avatars focus on recovering 3D human bodies from images or videos without the requirement for animation.
A class of methods~\cite{lin2023im4d,xu20244k4d,jiang2024hifi4g,jiang2024robust,zheng2024gps,zhou2024gps} employs multi-view image-based rendering strategies, using reference views to enhance the rendering quality from source views.
We compare our method with GPS-Gaussian~\cite{zheng2024gps}, one of the cutting-edge rendering-based method.
\cref{fig: gps} and~\cref{tab: gps} reveal that although GPS-Gaussian does not require per-scene training, it falls short of our method in novel-view rendering quality and lacks the ability to animate.

\begin{figure*}[h]
  \centering
  \includegraphics[width=1.\linewidth]{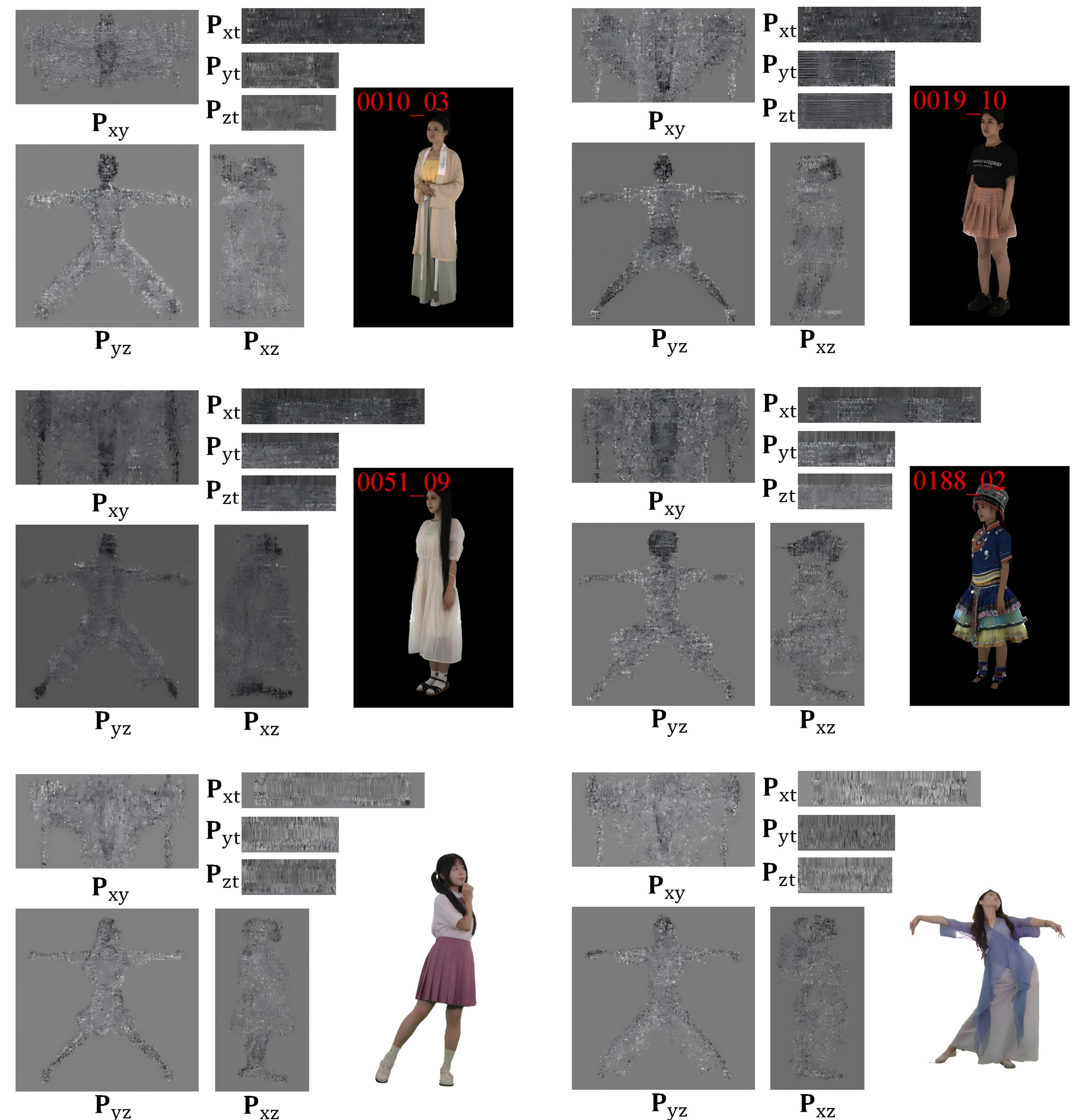}
  \caption{Hex-plane spatio-temporal encoder for different human avatars.}
  \label{fig: hexplanes}
\end{figure*}

\subsection{Partitioning of Human Body}
\label{subsec: Partitioning of Human Body}

\begin{table}[t]
\small
\begin{center}
\resizebox{.7\linewidth}{!}{
\begin{tabular}{lccc}
\toprule
& $\text{PSNR}\uparrow$          
& $\text{SSIM}\uparrow$          
& $\text{LPIPS}\downarrow$
\\
\cmidrule(r){1-4}
GPS-Gussian~\cite{zheng2024gps}
& 31.02 & 0.970  & 29.6 \\
$R^3$-Avatar (ours)
& \textbf{32.43}  & \textbf{0.974} & \textbf{25.2} \\
\bottomrule
\end{tabular}}
\end{center}
\vspace{-3ex}
\caption{
    \label{tab: gps} {Novel-view rendering comparison of our method and GPS-Gaussian on DNA-Rendering dataset~\cite{cheng2023dna}. LPIPS is $1000\times$.
    }
}
\end{table}

In Sec. 4.3. (main text), we propose a body partitioning strategy to retrieve timestamps for different body parts separately.
Our partitioning is based on the observed coupling of body motion.
For example, hands and legs typically do not affect the same region and are therefore separated, whereas the hip and waist are usually correlated and thus should be grouped together.
In~\cref{fig: body_seg} we display the results of partitioning on both SMPL~\cite{loper2015smpl} and SMPL-X~\cite{SMPL-X:2019}.

\begin{figure*}[t]
  \centering
  \includegraphics[width=.88\linewidth]{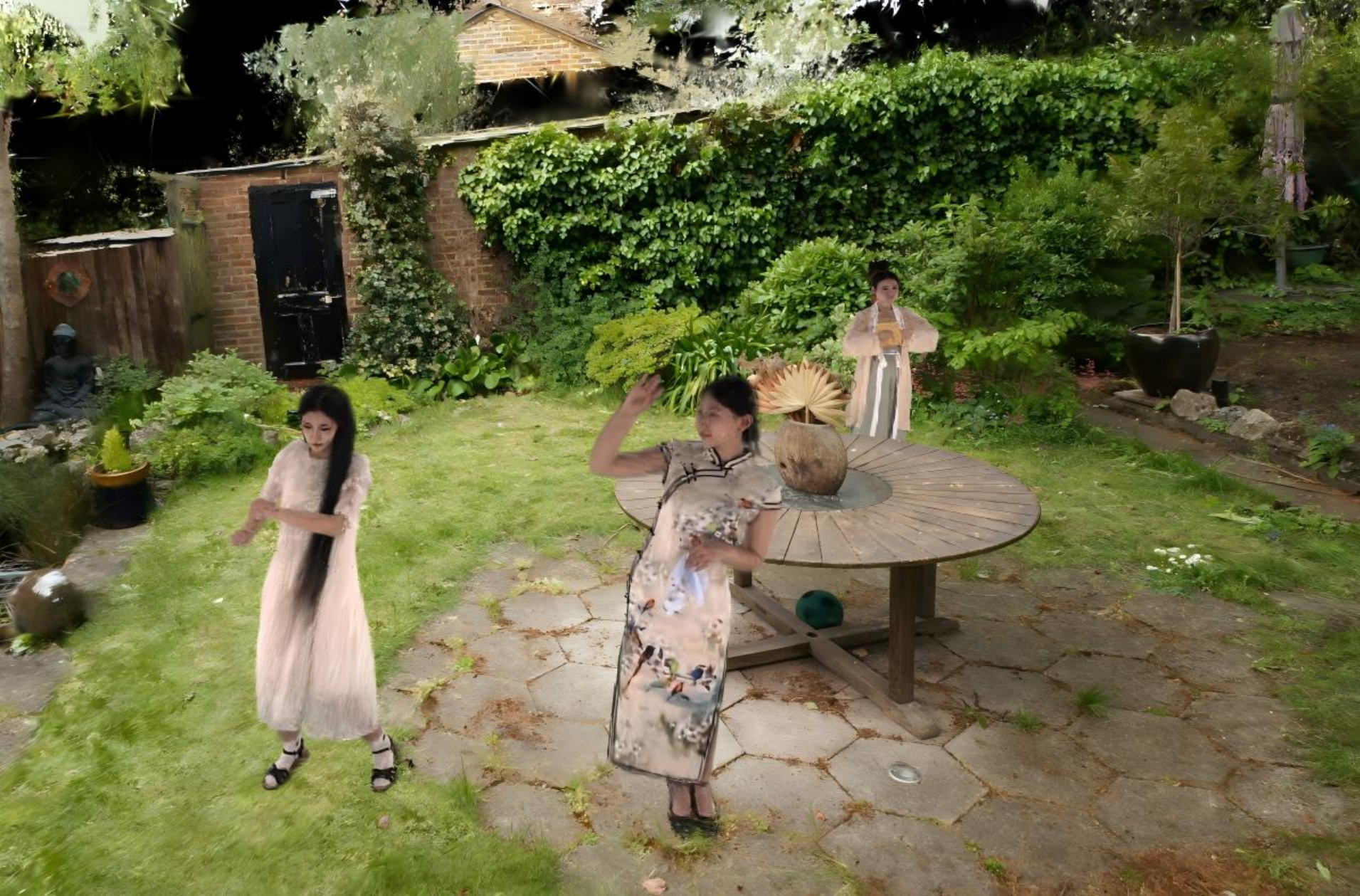}
  \caption{Render the human assets together with the scene.}
  \label{fig: scene}
\end{figure*}

\begin{figure*}[t]
  \centering
  \includegraphics[width=.88\linewidth]{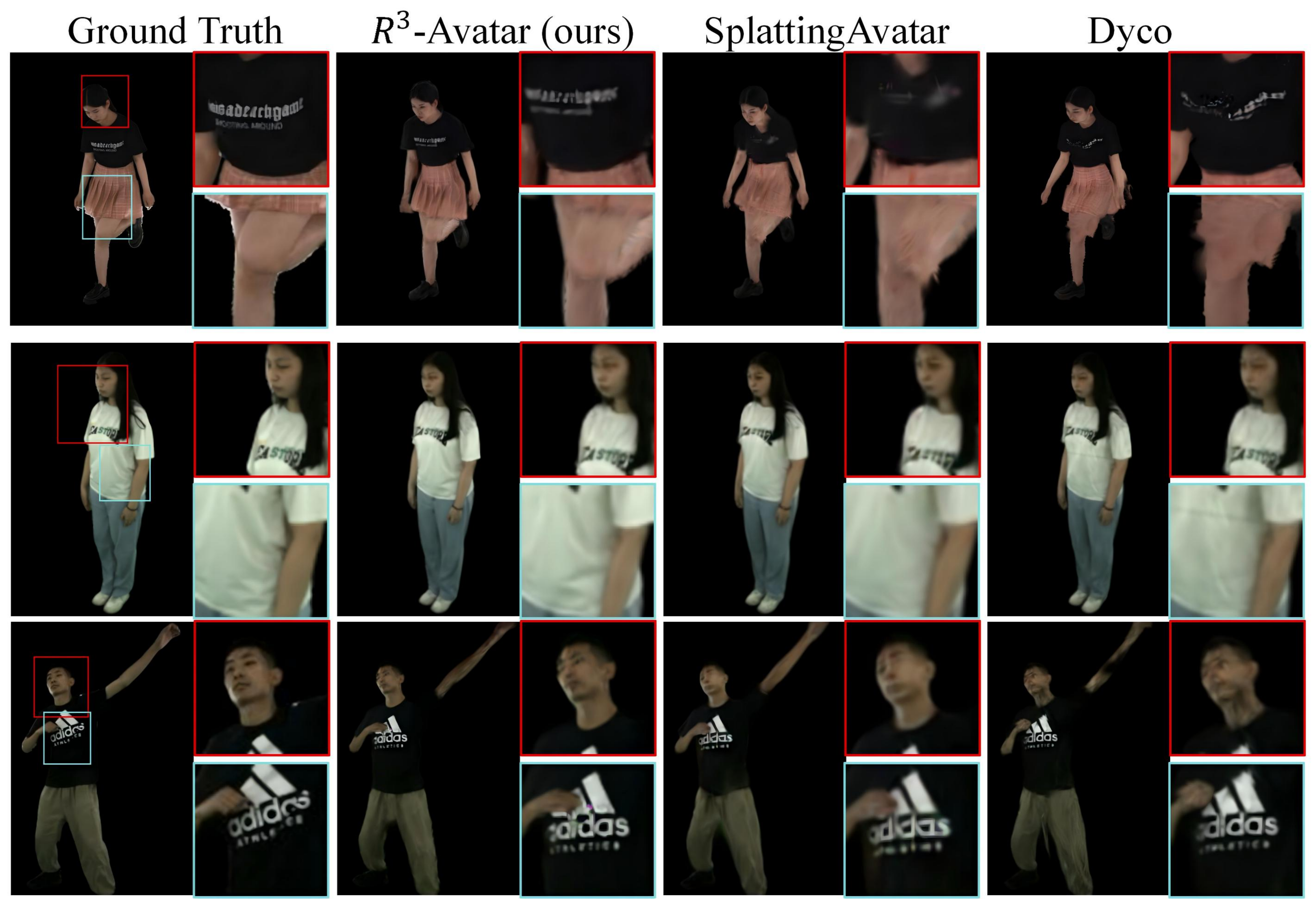}
  \caption{More results of novel-pose animating.}
  \label{fig: more_novel_pose}
\end{figure*}

\begin{figure*}[t]
  \centering
  \includegraphics[width=1.\linewidth]{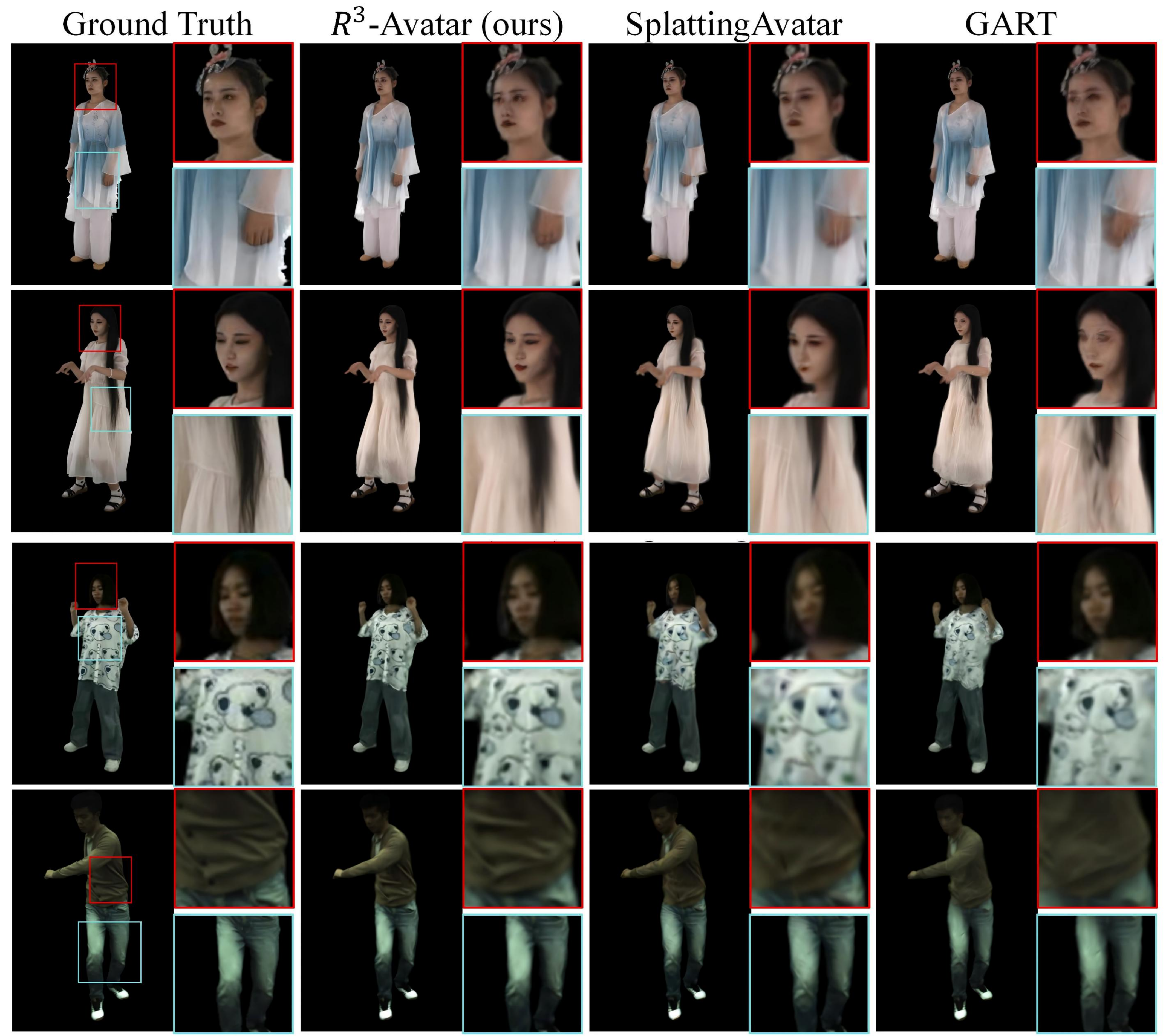}
  \caption{More results of novel-view rendering.}
  \label{fig: more_novel_view}
\end{figure*}

\subsection{Training and Rendering Efficiency}
\label{subsec: Training and Rendering Efficiency}

\begin{table}[t]
\small
\begin{center}
\resizebox{1\linewidth}{!}{
\begin{tabular}{lccccc}
\toprule
& 4DGS   & Dyco    & GART & SplattingAvatar & $R^3$-Avatar (ours) \\
\cmidrule(r){1-6}
Training Time
& 3h  & 16h  & 30min & 60min & 90min \\
Rendering Speed (FPS)
& 60+  & 0.5 & 60+ & 50+ & 60+ \\
\bottomrule
\end{tabular}}
\end{center}
\vspace{-3ex}
\caption{
    \label{tab: time efficiency} {Average training (till convergence) time and rendering speed on DNA-Rendering dataset~\cite{cheng2023dna}.
    }
}
\end{table}

We train all baselines (4DGS~\cite{yang2023real}, Dyco~\cite{chen2024within}, GART~\cite{lei2024gart}, SplattingAvatar~\cite{shao2024splattingavatar} and $R^3$-Avatar) on a single GeForce RTX3090.
Training time and rendering speed on DNA-Rendering dataset~\cite{cheng2023dna} are reported in~\cref{tab: time efficiency}.
Our method achieves better rendering and animation quality at a comparable rendering speed compared to GART and SplattingAvatar.
Dyco's training time and inference speed make it challenging to apply in real-time scenarios.

\subsection{Ablations on Hyperparameters of Recording Phase}
\label{subsec: Ablations on Hyperparameters of Recording Phase}

For hyperparameters of the spatial encoder, we apply multi-scale planes~\cite{fridovich2023k} with $3$ different resolutions at $64^2$, $128^2$, $256^2$. 
For hyperparameters of the temporal codebook, we only apply multi-scale encoding on spatial resolutions, keeping the time resolution to be $50$.
Now we have the biggest plane set ($\textbf{P}_{xy}[256\times256]$, $\textbf{P}_{xz}[256\times256]$, $\textbf{P}_{yz}[256\times256]$, $\textbf{P}_{xt}[256\times50]$, $\textbf{P}_{yt}[256\times50]$, $\textbf{P}_{zt}[256\times50]$), which is displayed in~\cref{fig: hexplanes}.
And the smallest plane set is ($\textbf{P}_{xy}[64\times64]$, $\textbf{P}_{xz}[64\times64]$, $\textbf{P}_{yz}[64\times64]$, $\textbf{P}_{xt}[64\times50]$, $\textbf{P}_{yt}[64\times50]$, $\textbf{P}_{zt}[64\times50]$).
The per-plane and per-scale features are multiplied across planes and concatenated across scales.
The per-plane and per-scale features ($\bm{f}_{xy}$, $\bm{f}_{xz}$, $\bm{f}_{yz}$, $\bm{f}_{xt}$, $\bm{f}_{yt}$, $\bm{f}_{zt}$) all have dimension $32$, so the final encoded feature $\bm{f}$ has dimension $128$.
For hyperparameters of 4D gaussian decoder and LBS network, we detail the number of layers and the width of the MLP.
$\Phi_{gd}$ has $D=2$ and $W=256$.
$\Phi_{\text{lbs}}$ has $D=4$ and $W=128$.

We conduct additional ablation experiments on hyperparameter settings. of 
First, we explore the impact of selecting different resolutions for the hex-plane encoder.
In~\cref{tab: ablation on the resolution of hex-plane encoder}, we test with four different spatial resolutions while keeping temporal resolutions to $50$ (``level N, Res'' means using ``N'' level of multi-scale planes with max resolution ``Res'').
We observe a performance improvement with increased resolution, with the marginal gains gradually diminishing.
As a trade-off, we select the resolution as $256$, striking a balance between minimizing memory consumption and satisfactory performance.

\subsection{Ablations on Hyperparameters of Retrieving Phase}
\label{subsec: Ablations on Hyperparameters of Retrieving Phase}
%此处需要消融retrieving部分，seq长度，window大小，分块多少，平均最近几个时间

For retrieving phase, length of delta pose sequence $S^N_{\Delta{\bm{p}}}$ is $2$, $k$ in $\mathcal{S}_{\text{top$k$}}$ is set to $20$, sliding window size $W$ is 3.

The ``delta pose sequence'' provides motion direction during retrieving, facilitating pose matching.
However, an excessive length of $S^N_{\Delta{\bm{p}}}$ will reduces matching samples in training sets.
We ablate the length of delta pose sequence in~\cref{tab: ablation retrieving}.

$\mathcal{S}_{\text{top$k$}}$ is the set of training poses (or delta pose sequences) ranked by similarity based on their L2 distance to the current novel pose (or delta pose sequence).
The sliding window prioritizes poses from this set, considering only those whose timestamps fall within the sliding window relative to the previous novel-pose frames (\eg, if the previous novel pose retrieved timestamp 30, then the timestamps within the sliding window range would be $[30 - W, 30 + W]$. Any pose within this range in $\mathcal{S}_{\text{top$k$}}$ will be prioritized).
If $k$ is too big, the retrieved pose is less reliable because more low-confidence poses have been included.
However, if $k$ is too small, it is likely that no poses within the sliding window will be found, increasing temporal jitters.
As for size of sliding window, a value too small reduces the number of poses falling within the sliding window, leading to temporal jitters, while a value too large diminishes the smoothing effect of the window, as even poses from distant timestamps may fall within the same window.
In~\cref{tab: ablation retrieving}, we also ablate these two items.

\begin{table}[t]
\small
\begin{center}
\resizebox{0.7\linewidth}{!}{
\begin{tabular}{lccc}
\toprule
Resolution
& $\text{PSNR}\uparrow$   
& $\text{SSIM}\uparrow$    
& $\text{LPIPS}\downarrow$\\
\cmidrule(r){1-1} 
\cmidrule(r){2-4}
level 1, 64
& 29.13
& 0.957
& 43.7 \\
level 2, 128
& 32.01
& 0.970
& 27.4 \\
level 3, 256 (ours)
& \textbf{32.43}
& \textbf{0.974}
& \textbf{25.2} \\
level 4, 512
& 32.47
& 0.974
& 25.0 \\
\bottomrule
\end{tabular}}
\end{center}
\vspace{-3ex}
\caption{
    \label{tab: ablation on the resolution of hex-plane encoder}{Ablations on the resolution of hex-plane encoder. We show the novel-view rendering results on DNA-Rendering dataset~\cite{cheng2023dna}. LPIPS is $1000\times$.}
}
\end{table}

\begin{table}[t]
\small
\begin{center}
\resizebox{0.8\linewidth}{!}{
\begin{tabular}{lcccc}
\toprule
Resolution
& $\text{PSNR}\uparrow$   
& $\text{SSIM}\uparrow$    
& $\text{LPIPS}\downarrow$
& $\text{FID}\downarrow$\\
\cmidrule(r){1-1} 
\cmidrule(r){2-5}
$|S^N_{\Delta{\bm{p}}}|=0$
& 24.94
& 0.942
& 49.9
& 3.45\\
$|S^N_{\Delta{\bm{p}}}|=2$ (ours)
& \textbf{25.21}
& \textbf{0.945}
& \textbf{47.2}    
& \textbf{3.39} \\
$|S^N_{\Delta{\bm{p}}}|=4$
& 24.73
& 0.941
& 49.7
& 3.42\\
\cmidrule(r){1-1} 
\cmidrule(r){2-5}
$k=5$
& 25.09
& 0.944
& 48.1
& 3.43\\
$k=20$ (ours)
& \textbf{25.21}
& \textbf{0.945}
& \textbf{47.2}    
& \textbf{3.39} \\
$k=50$
& 24.92
& 0.941
& 48.6
& 3.44\\
\cmidrule(r){1-1} 
\cmidrule(r){2-5}
$W=1$
& 25.11
& 0.944
& 47.6
& 3.44\\
$W=3$ (ours)
& \textbf{25.21}
& \textbf{0.945}
& \textbf{47.2}    
& \textbf{3.39} \\
$W=5$
& 25.03
& 0.942
& 48.1
& 3.46\\
\bottomrule
\end{tabular}}
\end{center}
\vspace{-3ex}
\caption{
    \label{tab: ablation retrieving}{Ablations on hyperparameters of retrieving phase. We show the novel-pose animation results on DNA-Rendering dataset~\cite{cheng2023dna}. LPIPS is $1000\times$.}
}
\end{table}

\section{More visual Results}
\label{subsec: More visual Results}

We display more visulization in~\cref{fig: more_novel_view} (novel-view rendering ) and~\cref{fig: more_novel_pose} (novel-pose animating).

\subsection{Human in the Scene}
\label{subsec: Human in the Scene}

Since $R^3$-Avatar employs 3DGS-based representation, the reconstructed avatar supports seamless integration with other GS assets, \eg, the scene, and is therefore compatible with conventional rendering pipeline for 3DGS.
\cref{fig: scene} presents an example of 3D human avatars integrated into a scene reconstructed by 3DGS.

\subsection{Supplemental Video}
\label{subsec: Supplemental Video}
We have also attached a supplementary video alongside this PDF file,
which consists of novel-view rendering and novel-pose animation, together with visual comparisons with GART~\cite{lei2024gart}, one of the current state-of-the-art methods.
Since images may not fully convey the effectiveness of human reconstruction, especially for the results of animation, we encourage the readers to watch the supplemental video.

\end{document}